\documentclass[journal, a4paper]{IEEEtran}

\usepackage[pdftex]{graphicx}
\usepackage{graphicx}
\usepackage{multicol}
\usepackage{multirow}
\usepackage{blindtext}
\usepackage{amsmath}
\usepackage{amssymb}
\usepackage{fancyhdr}
\usepackage{cite}
\usepackage[size=smaller]{caption}
\usepackage{booktabs} 
\usepackage{tikz}
\usepackage[utf8]{inputenc}
\usepackage{pgfplots}
\usetikzlibrary{plotmarks}
\usepackage{pdflscape}
\pgfplotsset{compat=newest}
\pgfplotsset{plot coordinates/math parser=false}
\newlength\fwidth 
\usepackage{enumitem} 
\usepackage{algorithmic}
\usepackage[ruled,vlined]{algorithm2e}
\usepackage{textcomp}
\usepackage{subcaption}
\usepackage{rotating}

\begin{document}

\title{SpikeFI: A Fault Injection Framework for \\ Spiking Neural Networks}

\author{
\IEEEauthorblockN{Theofilos Spyrou\IEEEauthorrefmark{1}, Said Hamdioui\IEEEauthorrefmark{1} and Haralampos-G. Stratigopoulos\IEEEauthorrefmark{2}\\}
\IEEEauthorblockA{\IEEEauthorrefmark{1}\small{Computer Engineering Lab, Delft University of Technology, Delft, The Netherlands\\}}
\IEEEauthorblockA{\IEEEauthorrefmark{2}\small{Sorbonne Universit\'{e}, CNRS, LIP6, Paris, France}}
}

\maketitle

\begin{abstract}

Neuromorphic computing and spiking neural networks (SNNs) are gaining traction across various artificial intelligence (AI) tasks thanks to their potential for efficient energy usage and faster computation speed. This comparative advantage comes from mimicking the structure, function, and efficiency of the biological brain, which arguably is the most brilliant and green computing machine. As SNNs are eventually deployed on a hardware processor, the reliability of the application in light of hardware-level faults becomes a concern, especially for safety- and mission-critical applications. In this work, we propose \textit{SpikeFI}, a fault injection framework for SNNs that can be used for automating the reliability analysis and test generation. \textit{SpikeFI} is built upon the SLAYER PyTorch framework with fault injection experiments accelerated on a single or multiple GPUs. It has a comprehensive integrated neuron and synapse fault model library, in accordance to the literature in the domain, which is extendable by the user if needed. It supports: single and multiple faults; permanent and transient faults; specified, random layer-wise, and random network-wise fault locations; and pre-, during, and post-training fault injection. It also offers several optimization speedups  and built-in functions for results visualization. \textit{SpikeFI} is open-source and available for download via GitHub at {\textcolor{blue}{\underline{https://github.com/SpikeFI}}}.

\end{abstract}


\section{Introduction}
\label{sec:intro}

Neuromorphic computing is an emerging computing paradigm that has its roots in mimicking the spike-based operation of neurons in the biological brain.
A neuromorphic processor essentially maps a Spiking Neural Network (SNN). SNNs can offer orders of magnitude more energy efficiency and inference speed compared to the more conventional Artificial Neural Networks (ANNs) \cite{RoJaPa19, SKPMDK22}. For this reason, SNNs open exciting new possibilities for realizing the next-generation Artificial Intelligence (AI) systems and for powering intelligent and autonomous edge devices with local AI processing. A major leap forward in the recent years is the development of several large-scale neuromorphic processors, e.g., SpiNNaker \cite{FGTP14}, TrueNorth  \cite{Merolla14}, Loihi \cite{Loihi18}, BrainScaleS \cite{BrainScaleS}, and Neurogrid \cite{Neurogrid}, supported also with software frameworks.

In this work, we address the dependability aspects of neuromorphic processors in view of the rare yet inevitable hardware-level faults. Hardware-level faults include bit-flips caused by cosmic ray particle strikes (a.k.a. soft errors) and defects and process parameter variations that are induced during manufacturing or occur in the field due to silicon aging mechanisms. SNNs show a large degree of inherent fault tolerance thanks to the analogy to the biological brain that has remarkable fault tolerance capabilities. Thus, most faults end up being benign: they are masked, i.e., their effect is not propagated to the output, or they can be tolerated, i.e., the output changes but the cognitive decision is still correct. However, there exist critical faults that will cause a wrong output disrupting the application.

More specifically, we propose a generic Fault Injection (FI) tool, named \textit{SpikeFI}, for automating fault analysis of SNNs. Starting with an SNN model, the user is able to inject faults on different locations in the SNN architecture and assess their impact on the success of training and the accuracy of inference. The tool supports any SNN model, i.e., fully-connected. convolutional, or recurrent. It embeds a comprehensive fault model library that can be customized by the user. It supports single or multiple faults, transient or permanent faults, as well as statistical fault injection layer-wise and network-wise. Fault injection can be performed before, during or after training. \textit{SpikeFI} also offers several simulation speedup options, such as early stop and late start, and has built-in various types of results visualisation functions.

\textit{SpikeFI} has several use cases:

\begin{enumerate}[leftmargin=*]
    \item Understand the vulnerability of the SNN application to faults. 
    \item Assess how architectural choices (i.e., depth, layer size, feature map size, weight quantization, network compression, etc.) and the different per-layer hyper-parameters (i.e., neuron threshold, leakage, and refractory period) affect the resilience to faults so as to make early design decisions with reliability in mind.
    \item Guide test generation algorithms aiming at generating test inputs for sensitizing and detecting critical faults \cite{TCWL21, CCHC-ML22, E-SSCS23,SpSt23}. Compact test sets can be used for post-manufacturing testing or can be replayed in idle times or periodically for in-field on-line testing.
    \item Evaluate training algorithms in terms of their fault tolerance capabilities and develop fault-aware training algorithms \cite{SPMJPKDP20}. For example, faults can be inserted during training, transiently across the epochs, to increase the robustness of the network to faults once deployed. Once the training is over the faults are ejected.
    \item Assess the criticality of faults towards developing cost-effective hardware-level fault tolerance techniques \cite{HBTL11,KHMcDGLHTTMJ17,JLMKTHTMH18, LHMMcDW18, E-SC-ML-BS19, SE-SAC-ML-BS21, PuHaSh22,SaAmCh23}. 
\end{enumerate}

\textit{SpikeFI} performs fault analysis at the application level in software. Fault injection can be performed instead at the hardware description level, i.e., RTL, gate-level or transistor-level, but this requires the availability of the hardware implementation and the more detailed the hardware description is, the lengthier the simulation time is. Already at RTL the simulation becomes intractable for sizeable SNN models given that the fault space explodes and that the impact of each fault is evaluated by performing inference on the complete testing set. Fault injection can also be performed on the actual hardware \cite{SE-SAC-ML-BS22}, but at this stage it may be too late to make any architectural changes for implementing fault tolerance techniques. To this end, \textit{SpikeFI} adopts the flexibility and speed of software-level fault injection while supporting hardware-aware fault models by mapping them to software operators and accelerating faulty SNN instances on a GPU.

\textit{SpikeFI} is publicly available and downloadable at \underline{https://github.com/SpikeFI}. It is open-source and extendable, allowing researchers to implement their own fault models and results analysis.

There exist several works that have employed custom-made FI frameworks for SNNs (for example, see \cite{VaDiNaAn19,SPMJPKDP20, TCWL21,SE-SAC-ML-BS21, E-SSCS23, PuHaSh22}), but none of these FI frameworks was made publicly available and open-source. Very recently, the \textit{SpikingJet} fault injector for SNNs was made publicly available \cite{GMCRSC24}. \textit{SpikingJet} is built on top of the SnnTorch framework \cite{SnnTorch}, whereas \textit{SpikeFI} is built on top of the Spike Layer Error Reassignment in Time (SLAYER) framework \cite{shor18}. Compared to \textit{SpikingJet}, \textit{SpikeFI} provides several simulation speedup options, it supports in addition transient faults and fault injection before training, and it offers results visualization. 

Software-level FI frameworks have also been developed for ANNs \cite{LHSTPEK17,RGPWLMBW18,MANVAFFH20,CNFLPD20, NCFLPD23}. Recent efforts aim at improving the one-to-one mapping between hardware and software fault injection \cite{HeBaLi20, BCMT23}, reproducing more complex fault models, i.e., extracted from radiation tests \cite{LRSCKSBD22},  or speeding up the analysis by reducing the fault injection space \cite{CLPD19,CTSC22} or the fault simulation time \cite{GaRuSa23,RGSGSSoReSMAA23}. Another possibility is to use generic FI tools \cite{HTSKE17, THSVK21} to emulate fault effects in the hardware platform, i.e., GPU, running the application. Such FI frameworks are crucial towards the testability and dependability of AI hardware accelerators \cite{SuLiSt23,RSLDTB23,StSpRa23, ATRDJ24,Re24}.

The rest of the article is structured as follows. In Section \ref{sec:background}, we provide background information on SNNs. In Section \ref{sec:fault_modeling}, we discuss fault modelling for SNNs used by \textit{SpikeFI}. The \textit{SpikeFI} framework is presented in Section \ref{sec:framework}. The results are presented in Section \ref{sec:results}. Section \ref{sec:conclusions} concludes this article.

\section{Background Information on SNNs}
\label{sec:background}

\subsection{Principle of operation}

Neural network models are classified into three generations. The first generation was based on McCulloch-Pitts neurons, also referred to as perceptrons, which give a digital output. The second generation of models applied an activation function to the output of the neurons, such as a rectified linear unit (ReLU) or sigmoid, and, in this way, they supported analog computation and learning algorithms based on gradient-descend, such as backpropagation. SNNs are the third generation \cite{Ma97}, distinguishing themselves from their predecessors by their ability to mimic more realistically the biological brain. However, they constitute simplified models of their complex biological counterparts maintaining some of their aspects, since they are primarily used for computational purposes, rather than simulating the human brain.

Inspired from biological neural systems, SNNs encode the information in the timing of single action potentials, or spikes, and incorporate the time between successive spikes as a source of computation and communication among their spiking neurons. Spikes correspond to events generated whenever a change occurs providing a continuous time processing with very detailed time resolution reaching the micro- or nanosecond scale.

From a hardware perspective, SNNs form the basis of neuromorphic computing. Spiking neurons operate asynchronously to each other, as they are only utilized when an incoming spike stimulates them via their pre-synaptic connections. This makes SNNs event-based computation systems, which offers low latency and energy consumption compared to level-based ANNs. However, there are still challenges facing SNNs, such as the complexity of training. 

Another characteristic of SNNs is the input type, which needs to be in a spiking form as well, i.e., the network is fed with a continuous-time event flow instead of static frames. A natural way to achieve this is with a neuromorphic camera, also known as Dynamic Vision Sensor (DVS). A DVS resembles the retina of the human eye and is composed of pixel-neurons that react to changes in brightness. When a sufficient change has occurred to the brightness of a pixel-neuron, it generates a positive or negative event, depending on the polarity of the change.

\subsection{Spiking neuron models}
\label{sec:srm}

There exist several spiking neuron models, ranging from biologically detailed ones, such as the Hodgkin-Huxley, 
to simplified ones that are more computational efficient for large networks and hardware implementation while still incorporating the main neuronal dynamics, such as the Integrate \& Fire (I\&F) \cite{GKNP14}. As \textit{SpikeFI} is built on top of SLAYER \cite{shor18}, and given that SLAYER employs the Spike Response Model (SRM) which is a generalization of the I\&F model, herein we discuss in detail the SRM.

In the SRM, the state of the neuron at any given time is described by its membrane potential $u$. At its resting state, the membrane potential is set to a low value $u_{rest}$. The neuron integrates the incoming spikes from the synapses at its input and the membrane potential is increased or decreased according to the spike polarity. Once the potential reaches a certain threshold $\theta$, the neuron fires a spike, which is propagated to the next layer of neurons via the synapses connected to its output, and the neuron is reset to its resting state again. At the same time, the neuron is regulated to not fire again for a while. The minimum time in-between successive spikes is called refractory period. 

To mathematically express the above functionality, the SRM considers that the action of a neuron at any given time is a response to both the incoming activity and the neuron's own output. For this purpose, two response functions are used, namely, the synaptic kernel $\epsilon$ and the refractory kernel $\eta$. The synaptic kernel $\epsilon$ describes the effect of an incoming spike train on the membrane potential and distributes the effect of the most recent incoming spikes on future output spike values, hence introducing temporal dependency. The refractory kernel $\eta$ incorporates the effect of the neuron's own spike train onto its membrane potential. Two functions used to represent kernels $\epsilon$ and $\eta$ are:

\begin{equation}\label{eq:synaptic_kernel}
	\epsilon(s) = \frac{s}{\tau_s} \cdot e^{1 - \frac{s}{\tau_s}} \cdot H(s)
\end{equation}
\begin{equation}\label{eq:refractory_kernel}
	\eta(s') = -2\theta \cdot \frac{s'}{\tau_{ref}} \cdot e ^{1 - \frac{s'}{\tau_{ref}}} \cdot H(s'),
\end{equation}

\noindent where $H(\cdot)$ is the unit step function, $\tau_s$ is the membrane time constant, $\tau_{ref}$ is the refractory time constant, $s$ is the time passed since the last pre-synaptic spike at the input of the neuron, and $s'$ is the time passed since the last post-synaptic spike at the output of the neuron.

The input and output spike trains of the neuron are denoted by $S_i(t)$ and $S_o(t)$, respectively:

\begin{equation}\label{eq:spiking_input_output}
	\begin{split}
		S_i(t) & = \sum_f{\delta(t-t_i^f)}\\ 
		S_o(t) & = \sum_f{\delta(t-t_o^f)},  
	\end{split}
\end{equation}

\noindent where $\delta$ is the Kronecker delta function to denote a spike and $t_i^f$ and $t_o^f$ represent the $f$-th firing time at the $i$-th neuron input and neuron output, respectively. 

The membrane potential is expressed as:

\begin{equation}\label{eq:membrane_potential}
		u(t) = \sum_i{ \omega_i (\epsilon \ast S_i)(t)}  + (\eta \ast S_o)(t) + u_{rest}, 
\end{equation}

\noindent where $\omega_i$ is the weight of the synapse driving the $i$-th neuron input and $\ast$ is the convolution product.

Assuming that the neuron has fired $F$ spikes so far, a new spike $\delta(t-t_o^{F+1})$ is fired at time $t_o^{F+1}$ when the neuron's membrane potential reaches the threshold $\theta$ and is appended to the output spike train as follows:

\begin{equation}\label{eq:spiking_output}
	\begin{split}
		S_o(t) & = \sum_{f=1}^{F}{\delta(t-t_o^f)} + \delta(t-t_o^{F+1}) \\ 
		t_o^{F+1} & = \min{\{t: u(t) = \theta, t > t_o^F\}}.
	\end{split}
\end{equation}

\noindent Once this happens, the neuron's membrane potential is reset to its resting state:

\begin{equation}\label{membrane_reset}
	u(t_{o}^{F+1}) = u_{rest}.
\end{equation}

\begin{figure}[t]
\centering
    \begin{subfigure}{1\columnwidth} 
        \resizebox{\textwidth}{!}{\input{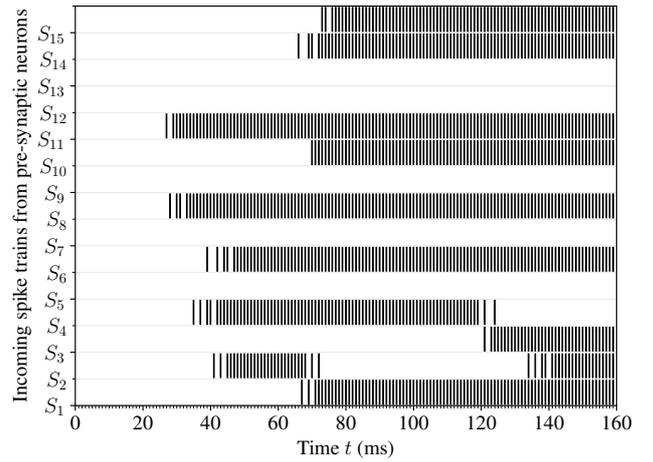}}
        \caption{Input spiking activity.}
        \label{fig:SRM_simulation_input}
    \end{subfigure}

    \begin{subfigure}{1\columnwidth} 
        \centering
        \resizebox{\textwidth}{!}{\input{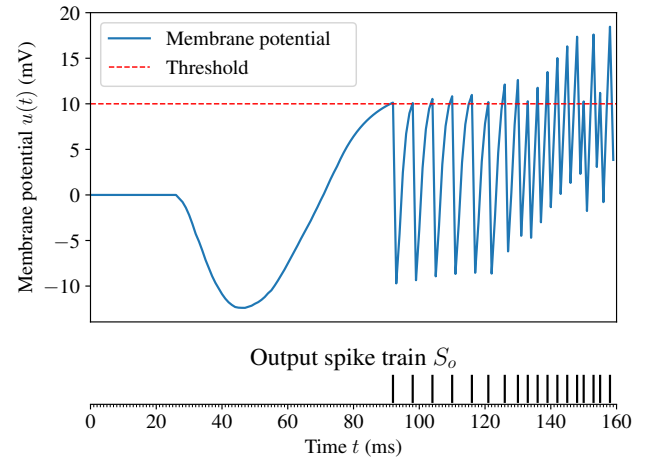}}
        \caption{Membrane potential and output spike train.}
        \label{fig:SRM_simulation_output}
    \end{subfigure}
    
    \caption{SRM simulation.}
    \label{fig:SRM_simulation}
\end{figure}

Fig. \ref{fig:SRM_simulation} shows a simulation of a neuron following the SRM. The neuron is receiving input from many neurons. The pre-synaptic spike trains from 20 out of these neurons are shown in Fig. \ref{fig:SRM_simulation_input}. Fig. \ref{fig:SRM_simulation_output} shows the evolution of the neuron's membrane potential and its output spike train.

\subsection{Training schemes}

Training large SNN models remains a challenge \cite{TGKMM19,RoJaPa19, SKPMDK22}. The classic backpropagation algorithm used in ANNs cannot be applied directly to work with spiking events due to their non-differentiable nature. Several learning schemes have been developed to overcome this challenge, including the biologically inspired Spike-Timing-Dependent Plasticity (STDP), training an ANN and converting it to a SNN, evolutional algorithms, and spike-based backpropagation. Spike-based backpropagation is currently one of the most practical and accurate techniques for training SNNs. SLAYER \cite{shor18} trains a SNN with a variation of backpropagation using the probability of a spiking neuron to change state, i.e., fire a spike or move back to its resting state. The fact that SNNs can have the same topologies as ANNs, e.g., fully-connected, convolutional, recurrent, etc., allows for embedding SLAYER in already mature Machine Learning (ML) frameworks with some adjustments to add support for the spiking functionality. To this end, there is a PyTorch version of SLAYER that can be used for both the training and the inference of any SNN model. \textit{SpikeFI} is built upon SLAYER and PyTorch, inheriting their features and extending their capabilities to support FI experiments.

\section{Fault Modeling}
\label{sec:fault_modeling}

\textit{SpikeFI} is a fault injector at the application level. Hardware-level faults are translated to behavioral-level faults which, thereafter, are reproduced mathematically into the SRM described in Section \ref{sec:srm}. We consider that the processing elements of the SNN, i.e., neurons and synapses, are discrete entities that can fail independently. A bottom-up approach can be followed to extract the spiking neuron and synapse faulty behaviors starting from transistor-level simulations \cite{E-SSAC-ML-BS20}. \textit{SpikeFI} conforms to the established practice and adopts all widespread and conventional fault models in the literature that are derived from various digital, analog, and mixed analog-digital SNN hardware implementations. These fault models are built-in within \textit{SpikeFI} and are delivered as a library. The library is fully modifiable and extendable, i.e., developers are free to create their own fault models depending on the hardware implementation and fault occurrence probabilities. 

Modeling faults at behavioral-level allows evaluating their impact without the need of knowing the details of their source or mechanism, avoiding in this way costly low-level simulations, i.e., at the transistor, gate, microarchitectural level, etc., performed at the network scale. Additionally, behavioral-level fault modeling provides the flexibility to model any possible hardware-level fault, as long as a mathematical formula describing it can be derived. Another advantage is that the results are not tied to a specific hardware accelerator design, thus the drawn conclusions tend to be more generic. 

Next, we describe the built-in fault models, which are summarized in Table \ref{tab:fault_models}, by separating them into neuron and synapse fault models. Fault models can be further divided into hard and parametric fault models, depending on whether the processing element presents an outright failure or deviation.

\begin{table}[t]
    \centering
    \caption{Behavioral-level representations of built-in fault models.}
    \begin{tabular}{cccc}
    \toprule        
        \multicolumn{2}{c}{Fault type}
        & {Fault model}
        & {Fault effect} \\ \toprule
        
        \multirow{6}*[-1.9em]{\rotatebox{90}{Neuron faults}}
        & \multirow{3}*[-0.75em]{\rotatebox{90}{Hard}}
        & Dead neuron
        & $\hat S_o(t) = 0$ \\ \cmidrule{3-4}
        
        && Saturated neuron
        & $\hat S_o(t) = \sum_{n = 0}^{\infty}{\delta(t - n)}$ \\ \cmidrule{3-4}
        
        && Stuck-at-x neuron
        & $\hat S_o(t) = x \cdot \sum_{n = 0}^{\infty}{\delta(t - n)}$ \\ \cmidrule{2-4}

        & \multirow{3}*[-0.2em]{\rotatebox{90}{Parametric}}
        & Integration fault
        & $\hat\tau_s=\rho \cdot \tau_s$ \\ \cmidrule{3-4}
     
        && Refractory fault
        & $\hat\tau_{ref}=\rho \cdot \tau_{ref}$ \\ \cmidrule{3-4}
        
        && Threshold fault 
        & $\hat\theta = \rho \cdot \theta$ \\ \midrule

\multirow{6}*[-1.9em]{\rotatebox{90}{Synapse faults}}
        & \multirow{2}*[-0.75em]{\rotatebox{90}{Hard}}
        & Dead synapse
        & $\hat\omega_i = 0$  \\ \cmidrule{3-4}
        
        && Saturated synapse
        & $\hat\omega_i >> \omega_i$ or $\hat\omega_i << \omega_i$ \\ \cmidrule{2-4}

        & \multirow{3}*[-0.2em]{\rotatebox{90}{Parametric}}
        & Perturbed synapse
        & $\hat\omega_i = \rho \cdot \omega_i$ \\ \\ \cmidrule{3-4}
        
        && Bit-flipped synapse 
        & $\hat\omega_i =  Q^{-1}( Q(\omega_i) \oplus 2^b )$\\
\\   
    \bottomrule
    \end{tabular}
    \label{tab:fault_models}
\end{table}

\subsection{Neuron faults}\label{sec:neuron_fault_modeling}

\subsubsection{Hard faults}

\paragraph{Dead neuron}

A fault that halts the neuron's spiking activity and makes it unresponsive to any input. To model a dead neuron, its output spike train is set to zero, i.e., $\hat S_o(t)=0$.

\paragraph{Saturated neuron}

A fault that causes a neuron to be firing non-stop, even in the absence of input activity. A saturated neuron can be considered as the complementary extreme case of a dead neuron, where the output spike train is never zero, i.e., $\hat S_o(t)=\sum_{n = 0}^{\infty}{\delta(t - n)}$.

\paragraph{Stuck-at-x neuron}

A fault that causes the neuron's output to be stuck-at a value $x$, i.e., $\hat S_o(t)=x \cdot \sum_{n = 0}^{\infty}{\delta(t - n)}$, where $x \in \mathbb{R}$. A stuck-at neuron can be viewed as the generic hard neuron fault, with the extreme dead and saturated neuron faults being derived by setting $x = 0$ and $x = 1$, respectively.

\subsubsection{Parametric faults}\label{sec:neuron_parametric_faults}

\paragraph{Integration fault}

A fault that causes the membrane time constant $\tau_s$ of the synaptic kernel $\epsilon$ in Eq. (\ref{eq:synaptic_kernel}) to be perturbed to a new value $\hat\tau_s=\rho \cdot \tau_s$, $\rho \in \mathbb{R}$, affecting the neuron's easiness to fire spikes. Depending on whether $\hat\tau_s > \tau_s$ or $\hat\tau_s < \tau_s$, the kernel function $\epsilon$ allows for an easier or harder integration, respectively, or, conceptually speaking, makes the neuron more or less sensitive towards incoming spikes at its input.

\paragraph{Refractory fault}

A fault that causes the refractory time constant $\tau_{ref}$ of the refractory kernel $\eta$ in Eq. (\ref{eq:refractory_kernel}) to be perturbed to a new value $\hat\tau_{ref}=\rho \cdot \tau_{ref}$, $\rho \in \mathbb{R}$, affecting the neuron's refractoriness, i.e., the minimum time that needs to elapse before the neuron is capable of firing again. If $\hat\tau_{ref} > \tau_{ref}$, then the refractory kernel $\eta$ converges slower to zero, meaning that the refractoriness of the neuron becomes stronger. On the other hand, if $\hat\tau_{ref} < \tau_{ref}$, then the neuron's state is less tightly associated to its own output activity and, thereby, it demonstrates a weak refractoriness.

\paragraph{Threshold fault}

A fault that causes the threshold of the neuron $\theta$ to be perturbed to a new value $\hat\theta = \rho \cdot \theta$, $\rho \in \mathbb{R}$. As it can be inferred from Eq. (\ref{eq:spiking_output}), a faulty threshold $\hat\theta$ affects the output spike train either in a contributory or in a suppressive way, depending on whether $\hat\theta < \theta$ or $\hat\theta > \theta$, respectively. Similarly to an integration fault, a lower threshold leads to a neuron that fires easier if excited by the same stimuli, as the membrane potential reaches the threshold faster. This fault has a second effect since the threshold is also used in the refractory kernel $\eta$. Therefore, from Eq. (\ref{eq:refractory_kernel}), an increase in the threshold's value, implies a higher refractory period, thus taking the neuron longer to recover after firing, while a decrease has the reverse effect.

\subsection{Synapse faults}\label{sec:synapse_fault_modeling}

\subsubsection{Hard Faults}

\paragraph{Dead synapse}

A fault that holds the synaptic weight to zero, i.e., $\hat{\omega_i} = 0$, disabling the synapse and ``cutting'' the connection between the pre- and post-synaptic neurons.

\paragraph{Saturated synapse}

A fault that saturates the synaptic weight to an extreme positive $\hat\omega_i >> \omega_i$ or extreme negative $\hat\omega_i << \omega_i$ value. Some representative positive and negative saturation values could be respectively the highest and lowest values of the weight distribution resulting from training.

\subsubsection{Parametric Faults}

\paragraph{Perturbed synapse}

A fault that perturbs the synaptic weight to a new value $\hat{\omega_i} = \rho \cdot \omega_i$, $\rho \in \mathbb{R}$.

\paragraph{Bit-flipped synapse}

From a hardware perspective, representing synaptic weights as digital words and storing them in on-die memories is common practice. This fault flips one or more bits of a $N$-bit representation of the synaptic weight. For example, for a $N$-bit integer representation, to model this type of fault, the real value of the synaptic weight is quantized with a precision of $N$ bits, the selected bits are flipped, and then the weight is converted back to a real number. At the end, the faulty value of the synaptic weight is given by $\hat\omega_i = Q^{-1}( Q(\omega_i) \oplus 2^b)$, where $Q$ is the quantization function, $Q^{-1}$ is the inverse quantization function, $\oplus$ is the logic XOR operator, and $b$ represents the position(s) of the flipped bit(s), where $b=0$ corresponds to the Least Significant Bit (LSB). This fault model can assume a Bit Error Rate (BER) probability for the memory storing the synaptic weights of the network in a binary data type.

\subsection{Permanent and transient fault effect}
\label{sec:fault_modeling_transient}

Depending on the nature of a fault and the factors that led to its occurrence, its effect may be either permanent or transient. SNNs have a global internal clock that defines the discrete times when neurons can be firing. If a fault is transient, its effect is active for a limited period or number of clock cycles. For the rest of the FI experiment time, the components affected by the transient fault are restored to their nominal state and, therefore, the original behavior is expected. For example, denoting the transient fault duration by $[t_1, t_2]$, the behavioral description of a transient neuron saturation fault is:

\begin{equation}\label{eq:neuron_saturation_transient_fault}
\hat S_{o}(t)= \left \{ \begin{array}{r@{\quad:\quad}l} \delta(0) & t \in [t_1, t_2] \\ S_o(t) & \mbox{otherwise}\end{array} \right.
\end{equation}

\noindent and the behavioral description of a perturbed synapse fault is given by:

\begin{equation}\label{eq:synapse_transient_fault}
\hat\omega_i (t)= \left \{ \begin{array}{r@{\quad:\quad}l} \rho \cdot \omega_i & t \in [t_1, t_2] \\ \omega_i & \mbox{otherwise}\end{array} \right..
\end{equation}

\section{The \textit{SpikeFI} Framework}
\label{sec:framework}

\subsection{Overview and supported features}

\textit{SpikeFI} is a native PyTorch framework \cite{Pytorch19} built upon the SLAYER framework \cite{shor18}, extending the capabilities of SLAYER to add support for setting and automatically executing FI and reliability analysis experiments. SLAYER is a notable training framework for SNNs that is contributing to the growing interest in SNNs. It is developed to enable efficient spike-based backpropagation learning for deep SNNs. It has been incorporated into Lava, Intel's open-source software framework for developing SNNs. \textit{SpikeFI} employs the same principles and programming concepts and paradigms as its underlying frameworks PyTorch and SLAYER preserving compatibility. Any arbitrary SNN model implemented in SLAYER can be the subject of FI and reliability analysis, without needing to make any modifications to the model. Researchers and developers already familiar with SLAYER can therefore jump directly into using \textit{SpikeFI}. \textit{SpikeFI} is offered as open-source software and is available for download and to contribute via the GitHub platform: \underline{github.com/SpikeFI/}.

\begin{figure}[t]
    \centering  \includegraphics[width=1\columnwidth]{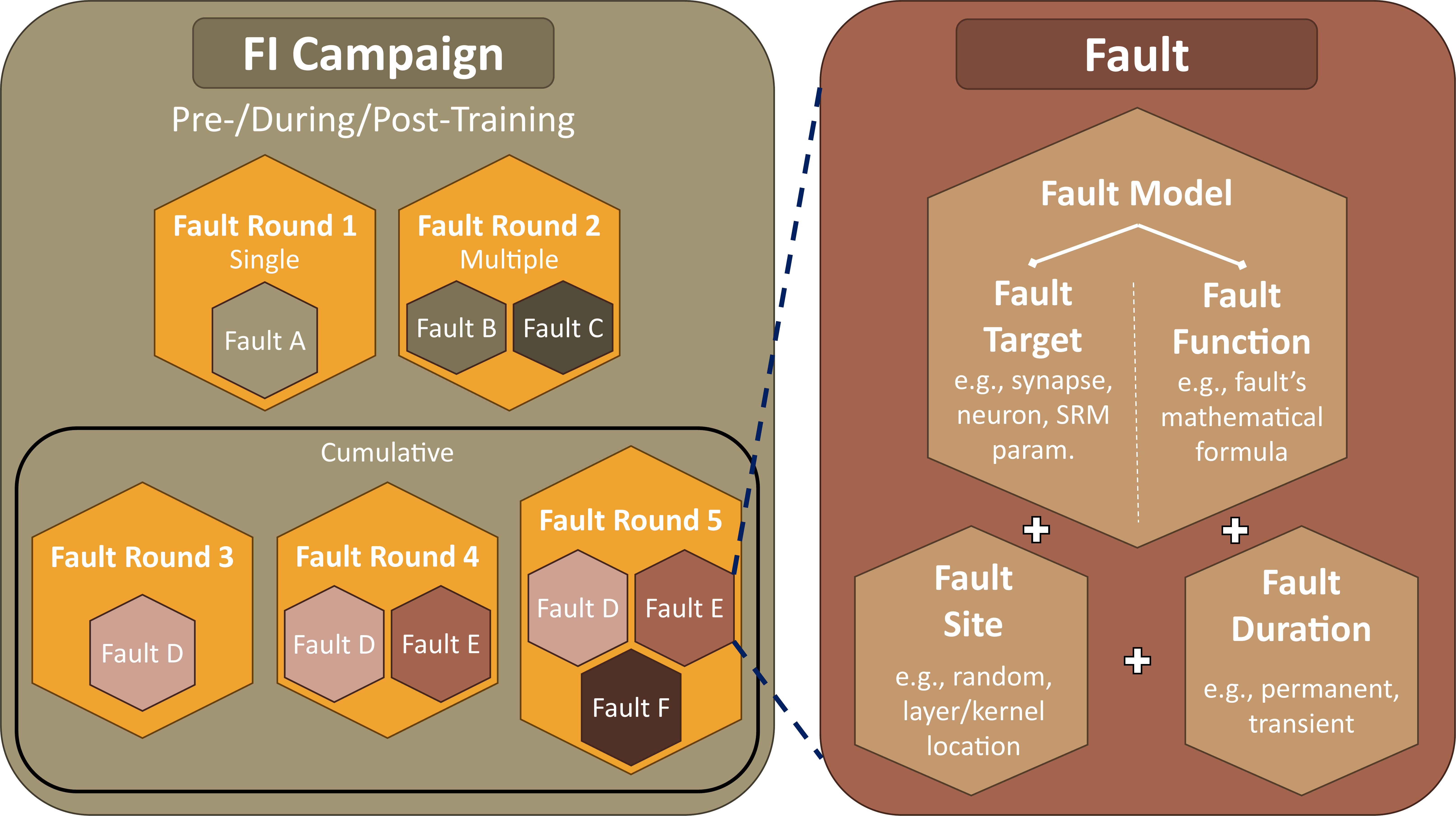}
    \caption{The organization of an FI campaign.}
    \label{fig:fi_organization}
\end{figure}

\textit{SpikeFI} supports the following features and scenarios:

\begin{enumerate}[leftmargin=*]
    \item \textit{Flexible fault modeling scheme}: \textit{SpikeFI} has an integrated comprehensive library of predefined fault models to select from, as described in Section \ref{sec:fault_modeling}. Faults can be injected into any processing element, i.e., neuron or synapse, and at different levels, i.e., isolated processing element, layer-wise, and network-wise. \textit{SpikeFI} being open-source allows the user to design and integrate custom fault models.
    \item \textit{Pre-/during/post-training FI injection}: \textit{SpikeFI} allows injecting faults before, during, or after the training phase prior to inference. The motivation is different across these scenarios. In pre-training fault injection, the goals can be to perform fault-aware training, re-train the network after the occurrence of a critical fault, and, in general, to study the network's capability to learn around faults. During training fault injection aims at studying the effect of faults occurring while training is progressing. This is useful when training deep SNNs as it can take significant time during which the hardware can suffer a fault. Post-training fault injection aims at studying the effect of faults on the inference accuracy. The analysis here aims at studying the inherent reliability and deriving critical fault types and locations. This information can subsequently be used for developing cost-effective test and fault tolerance strategies.
    \item \textit{Multi-round FI campaign}: A FI campaign may be composed of multiple experiments that are conducted sequentially independent of each other. 
    \item \textit{Single/multiple/cumulative FI injection}: Each of the fault rounds may contain a single fault, multiple faults or accumulated faults. In the latter scenario, in each FI experiment the set of faults is increased to observe the accumulative effect of the new faults added.
    \item \textit{Permanent/transient fault analysis}: A fault can be designed to be permanent or transient of varying duration, as described in Section \ref{sec:fault_modeling_transient}.
    \item \textit{Optimization options}: \textit{SpikeFI}, being built on top of PyTorch, utilizes GPU acceleration. It also offers various optimization options to speedup fault injection experiments, such as proper \textit{for}-loop ordering, late start, early stop, and batch-wise inference, which will be described in detail in Section \ref{sec:framework_optimizations}. 
    \item \textit{Results visualization}: \textit{SpikeFI} optionally saves spike trains for off-line analysis, but it also offers several built-in results visualization functions based on the inference accuracy drop metric, as it will be discussed in Section \ref{sec:results_visualization} and will be demonstrated in Section \ref{sec:demonstrations}.
\end{enumerate}

\subsection{Structuring of a FI experiment}

Fig. \ref{fig:fi_organization} summarizes the hierarchical organization of programming elements that shape a FI campaign in \textit{SpikeFI}. 

\subsubsection{Fault Round}
\label{sec:framework_structure_fr}

A group of faults to be injected altogether into the network. A fault round can contain a single or multiple faults. The collective effect of all faults belonging to the same round is evaluated simultaneously in a single inference. 
\textit{SpikeFI} also offers the possibility to perform cumulative fault analysis, i.e., define multiple fault rounds where each fault round contains the faults of the previous fault round plus some additional faults. Once a fault round has been evaluated, the faults are withdrawn from the network before continuing to the next fault round.

\subsubsection{Fault}\label{sec:framework_structure_f}

The actual fault to be injected into a network, composed of a fault model, one or more fault sites, and a fault duration. 

\subsubsection{Fault model}\label{sec:framework_structure_fm}

The fault model is in turn composed of a fault target and a fault function. 

\paragraph{Fault target}

The fault target can be: (i) the neuron output; (ii) the SRM parameters, i.e., neuron's membrane time constant, threshold, and refractory time constant; and (iii) the weight of a synapse, as described in Sections \ref{sec:neuron_fault_modeling} and \ref{sec:synapse_fault_modeling}. 

\paragraph{Fault function}

The fault function is the mathematical formula that describes how the fault is to affect the operation of the targeted processing element, as described in Sections \ref{sec:neuron_fault_modeling} and \ref{sec:synapse_fault_modeling}.

\subsubsection{Fault Site}

The fault site is the location of the fault within the network. The fault can be isolated affecting a specific processing element or it can be applied to multiple processing elements simultaneously. The user has also the option to create random fault sites per layer or across the network following some statistical fault distribution. The site of a fault is composed of the layer and coordinates of the processing element within the layer. In the case of a neuron, the site is the quadruplet $(l,c,x,y)$, where $l$ is the layer name or number, $c$ is the feature map number within the layer, and $(x,y)$ are the coordinates of the neuron within the feature map. In the case of a synapse, the site is defined by the quadruplets of the two neurons connected by the synapse.

\subsubsection{Fault Duration}

The fault duration provides information on when the fault is activated and for how long it remains active, as described in Section \ref{sec:fault_modeling_transient}. 
By default, a permanent fault means that is active for the whole duration of the input sample, whereas the duration of a transient fault is only a portion of the duration of the input sample.

\subsection{FI implementation into SLAYER}
\label{sec:FI_in_SLAYER}

\begin{figure}[t]
\centering
\includegraphics[width=1\columnwidth] {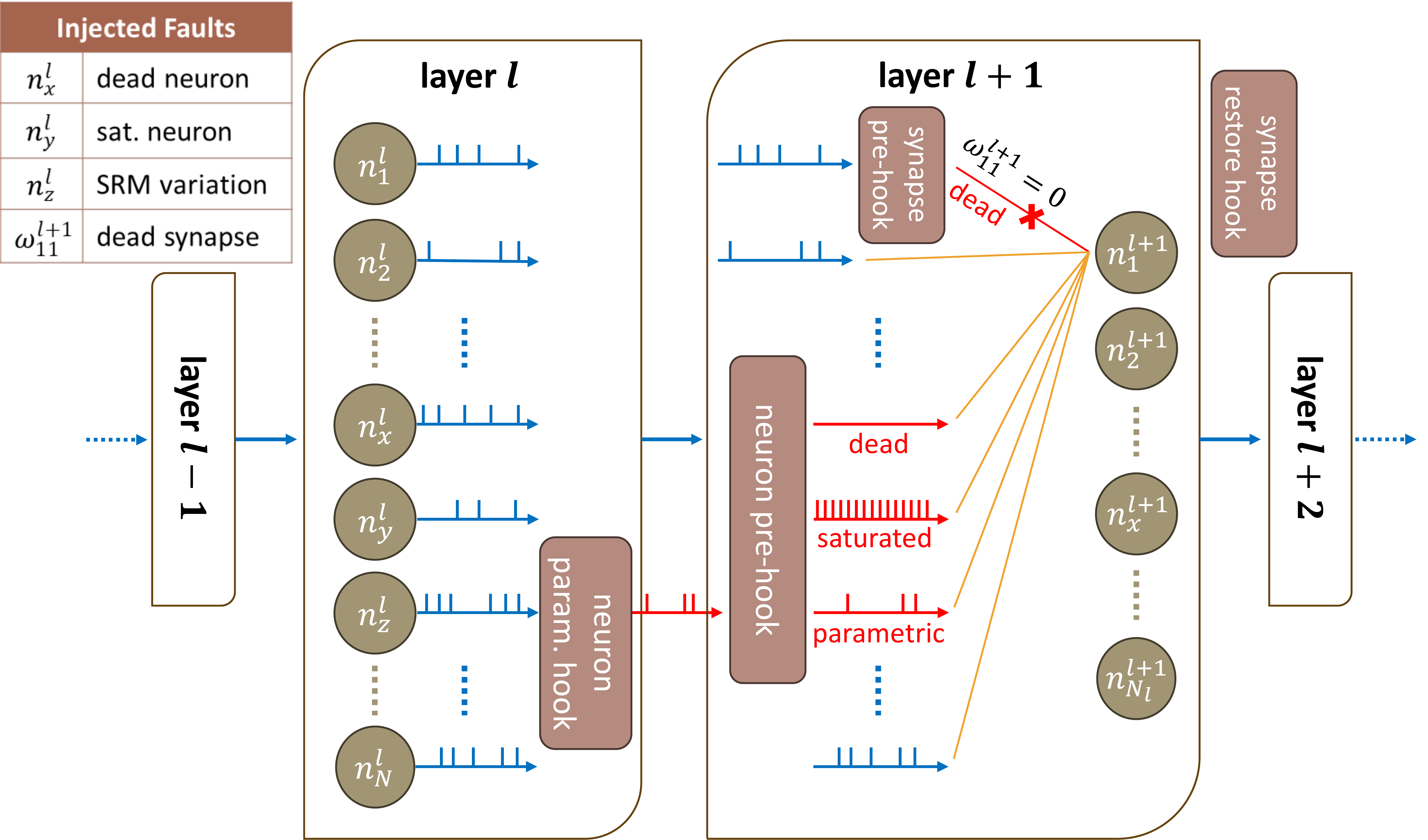}
    \caption{\centering FI implementation into SLAYER.}
    \label{fig:inj_methodology}
\end{figure}

To inject a fault, \textit{SpikeFI} modifies the PyTorch computation flow in SLAYER. For every fault model, there is a suitable modification to be applied, as illustrated in Fig. \ref{fig:inj_methodology}. To apply the modification, \textit{SpikeFI} makes use of PyTorch pre-hook and hook functions, i.e., a function called right before and right after the evaluation of a module, respectively.

Let us first consider neuron hard faults in layer $l$. The fault function of the fault model returns the output of the faulty neurons and updates the output of layer $l$ for these neurons while the rest of the neurons retain their fault-free output. Then, the output of layer $l$ is propagated to the input of the subsequent layer $l+1$ of the network. This is implemented with a neuron pre-hook function attached to the input of layer $l+1$, as illustrated in Fig. \ref{fig:inj_methodology}.

Regarding neuron parametric faults, in SLAYER the SRM parameters are set globally for a layer and are shared among its neurons, thus it is not possible to modify the parameters only for a subset of neurons. For this reason, \textit{SpikeFI} uses a hook and a pre-hook function. The first one, called neuron parametric hook, receives the same input as the faulty layer $l$, creates a ``dummy'' copy of the layer with altered SRM parameter for all neurons according to the fault model, repeats the parts of the calculation on this ``dummy'' copy that concern the affected SRM parameters, and feeds the result to the next neuron pre-hook function attached to the input of layer $l+1$. The neuron pre-hook function selects only the faulty neuron(s) and replaces their output spike train with the spike train of the corresponding neuron(s) in the dummy layer.

Regarding the synaptic weights, \textit{SpikeFI} uses a synapse pre-hook function to alter the synaptic weight value according to the fault model prior to the forward pass through the faulty layer. At the end of the faulty layer evaluation, there is a synapse restore hook function that restores the original synaptic weight, so that there is no interference between successive fault rounds.

\subsection{Optimizations}
\label{sec:framework_optimizations}

\subsubsection{Ordering of nested for loops}\label{sec:ordering_for_loops}

All fault rounds are evaluated by performing inference for the same set of input samples, which could correspond to the complete testing dataset or part of it. Essentially, a FI campaign is a nested loop iterating over all fault rounds and over the set of input samples. The ordering of the two \textit{for} loops has in fact an effect on the FI campaign runtime. \textit{SpikeFI} places the input samples in the outer \textit{for} loop and the fault rounds in the inner \textit{for} loop since this results in faster runtime, as it will be demonstrated quantitatively in Section \ref{sec:optimization_speedups}. The underlying reason is that the alternative \textit{for}-loops ordering would require transferring the dataset to the GPU for the computation multiple times, equal to the number of fault rounds, which would add a significant time overhead. Instead, with the selected \textit{for}-loop ordering, a batch transferred to the GPU is reused for all fault rounds, thus circumventing this time overhead.

\subsubsection{Late Start}\label{sec:late_start}

As typically there are multiple fault rounds in a FI campaign, there is significant repetition of forward passes through initial fault-free layers of the network. For example, consider two fault rounds with a single fault each, located at layers $i$ and $j>i$, respectively. The forward pass is repeated twice up to layer $i-1$ which is redundant. To save simulation time, for every fault round, \textit{SpikeFI} uses the late start option that skips layer computations up to the leftmost faulty layer, denoted by $l_{left}$. 

For this purpose, in a preparatory phase prior to the FI experiment, \textit{SpikeFI} performs a nominal inference for the complete testing dataset and records the deterministic golden output of all layers. More formally, for each layer $l$, \textit{SpikeFI} computes the matrix $A^l$ of its output spike trains with dimensions $N^l \times d$, where $N^l$ is the number of flattened neurons in layer $l$ and $d$ is the number of timestamps within the inference window. The SNN has a global clock with period $T$. The timestamps are denoted by $t_j=j*T$, $j=1,\cdots d$. $A^l(i,j)=1$ if neuron $i$ in layer $l$ fires at timestamp $t_j$, otherwise $A^l(i,j)=0$. The layer matrices $A^l$ are combined to generate the network matrix $A=[A^1, \cdots, A^L]$, where $L$ is the number of layers.

During the FI experiment, for every fault round, \textit{SpikeFI} first orders the faults based on the layer wherein they occur in ascending order. If the first faults appear at layer $l_{left}$, then \textit{SpikeFI} uses the golden output of layer $l_{left}-1$ and continues the simulation from this point onward. 

Note that if the fault round contains only hard neuron faults, then the simulation can continue from layer $l_{left}+1$. This is because of the implementation in PyTorch which injects the fault with a neuron pre-hook function attached to the input of the layer following the faulty layer. This means that late start can offer speeds-up even when the leftmost faulty layer is the first layer.

\subsubsection{Early Stop}

As mentioned in the introduction, many faults end up being benign. If in a fault round the output of the rightmost faulty layer, denoted by $l_{right}$, is unaffected matching the golden response, then it is pointless to continue the simulation as it will be like simulating a nominal fault-free network. \textit{SpikeFI} uses the early stop option to skip this redundant computation. More specifically, early stop uses the golden layer output matrices $A^l$ as in late start. Assuming a fault round with the last faults occurring in layer $l$, the early stop option computes the same matrix denoted by $A^l_{f}$, where the subscript $f$ indicates a fault round, subtracts it from the golden matrix $A^l$ to produce the matrix $B^l=A^l-A^l_{f}$, and computes the summation of all elements of $B^l$ denoted with the elementwise 1-norm $\| B^l \|_1$. If $\| B^l \|_1=0$ it means that all faults in this fault round are benign and the simulation stops at layer $l$. If $\| B^l \|_1 > 0$ it means that with respect to the golden output there is a spike count or spikes timing difference.

\textit{SpikeFI} also offers the possibility of considering a tolerance $\epsilon$ for the early stopping criterion. In this case, the simulation stops if $\| B^l \|_1 \leq \epsilon$. This tolerance should be used with care as it is likely that a fault induces a small spike count or spikes timing difference within the tolerance, yet this small difference is sufficient to cause the network's output spike trains to change to the point where the top-1 prediction changes. In this case, we stop early the simulation of a fault round that contains critical faults, mislabelling this fault round as benign.

\subsection{Complete FI experiment flow}

A FI experiment is divided into three stages, namely the preparation, execution, and results extraction and visualization stages.

\subsubsection{Preparation stage}\label{sec:framework_FI_campaign, preparation_stage}

First, the validity of all faults in all fault rounds is checked. If a fault is invalid, for example the fault site is nonexistent, then the fault is dropped from the experiment. Then, random faults are assigned a random site. After these two steps, the faults within each fault round are ordered according to the layer they belong to in ascending order so as to identify the leftmost $l_{left}$  and rightmost $l_{right}$ layers in the fault round. The verified fault rounds together with their sorted list of faults is communicated back to the user to acknowledge the FI experiment setup. Lastly, for each batch and before the FI starts, \textit{SpikeFI} performs an inference on the nominal network so as to record the golden outputs of all layers and form the matrices $A^l$ used in the late start and early stop optimizations.

\subsubsection{Execution stage}\label{sec:framework_FI_campaign_execution_stage}

\begin{figure*}[t]
    \centering
    \includegraphics[width=1\textwidth]{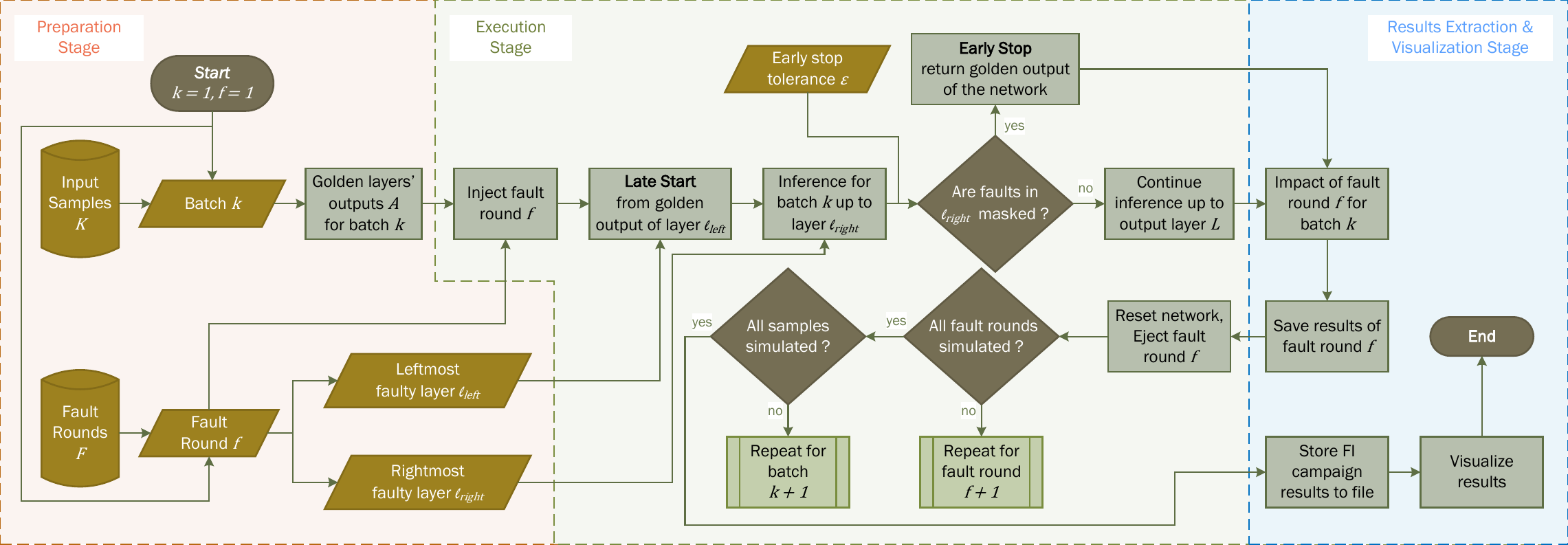}
    \caption{Flowchart of a FI campaign in \textit{SpikeFI}.}
    \label{fig:flowchart}
\end{figure*}

Fig. \ref{fig:flowchart} shows the \textit{SpikeFI} complete FI experiment flow including all optimizations. The dataset is transferred to the GPU in batches and for each batch the same FI experiment is performed in parallel for every input sample in the batch. The FI experiment is composed of a number of fault rounds simulated sequentially. For a given fault round, the simulation starts from the leftmost faulty layer $l_{left}$ using the golden output of the previous layer, or from layer $l_{left}+1$ if $l_{left}$ comprises only hard neuron faults. Inference continues up to the rightmost faulty layer $l_{right}$. In the case of a fault round with a single fault, $l_{left}$ and $l_{right}$ coincide. At this point, if the early stop criterion is met then the simulation stops and we proceed to the next fault round. Otherwise, the simulation continues up to the last layer. Before the next fault round starts, the results are saved and the network is initialized back to its original fault-free state.

\subsubsection{Results extraction and visualization stage}\label{sec:results_visualization}

Based on the spike encoding method employed by the SNN, i.e., rate coding, temporal coding, etc., for each fault round, \textit{SpikeFI} uses the output spike information in the matrices $A^L$ and $A^L_f$ to report the accuracy of the faulty network. For example, let us assume rate encoding, which is the most widely used spike encoding method, and a classification cognitive task. In this case, the output layer comprises one neuron per class and the winning class, i.e., the top-1 prediction, is that whose neuron fires the largest number of spikes within the inference time window. These spike counts are computed using the matrices $A^L$ and $A^L_f$ to assess if the faulty network changes the top-1 prediction of an input sample which contributes to accuracy loss. The user can define a misprediction tolerance value. If the accuracy drop is larger than this tolerance value, then this signifies a critical fault round. In this way, fault rounds are labeled as critical or benign. Optionally, the matrices $A^l$ and $A^l_f$, or their last parts $A^L$ and $A^L_f$, can be saved for offline analysis by the user. Finally, \textit{SpikeFI} reports the runtime to complete the FI campaign. 

\textit{SpikeFI} comes with several built-in results visualization functions that work for any fault model, as it will be shown in Section \ref{sec:results}. For example, misprediction rates can be plotted for isolated and random faults in the form of bar plots and heat maps and for parametric faults as a function of the parameter deviation. Results can also be presented bit-wise and layer-wise so as to perform comparisons.

\subsection{Open source code and example}
\label{sec:code_example}

A detailed documentation is provided in the GitHub platform for the easy integration and usage of the \textit{SpikeFI} framework. The structure of the \textit{SpikeFI} framework is organized in the form of a Python package to be imported and use its functions immediately. There are four main modules within the Python package, namely, the \textit{core}, \textit{fault}, \textit{models}, and \textit{visual} modules, each containing relative classes and functions to implement the framework's functionalities. An additional \textit{demo} module is included with the implementations of the two SNN architectures of Section \ref{sec:case_studies} and example scripts to showcase the usage of \textit{SpikeFI} in various scenarios.

\SetKwComment{Comment}{\# }{ }
\begin{algorithm}[t]
    \caption{Example of a FI campaign in \textit{SpikeFI}.}
    \label{alg:algorithm_example}
    \KwData{$net, shape\_in, slayer, test\_set$}

    $cmpn \gets Campaign(net, shape\_in, slayer)$ \\
    $f_x \gets Fault(DeadNeuron(), FaultSite(SF2))$ \\
    $f_y \gets Fault(SatuSynapse(10), FaultSite(SF1))$ \\
    $f_z \gets Fault(ParamNeuron(theta, 0.5), 4)$ \\
    $cmpn.inject(f_x)$ \\
    $cmpn.then\_inject(f_y, f_z)$ \\
    $cmpn.run(test\_set)$ \\
    $cmpn.export()$\\
    $cmpn.save()$ \\
    $bar(cmpn)$ \\
    $cmpn.eject()$ \\
    $cmpn.inject\_complete(BitflippedSynapse(7), SF2)$ \\
    $cmpn.run(test\_set)$ \\
    $heat(cmpn)$

\end{algorithm}

Algorithm \ref{alg:algorithm_example} presents a pseudo-code example resembling Python syntax of two FI campaigns. First, the campaign object $cmpn$ is created and initialized with the network model $net$, a vector with the dimensions of the input data samples $shape\_in$, and the spiking-related information object $slayer$, which is provided by SLAYER and is initialized by the user. More specifically, $slayer$ contains information about the SRM parameters of the spiking neurons, the duration of the input data samples, the global clock period, the target number of spikes for the winning class neuron, etc. Next, the faults $f_x$, $f_y$, $f_z$ are defined corresponding respectively to a dead neuron, a positively saturated synapse with value $10$, and a parametric neuron fault that decreases parameter $\theta$ to $50\%$ of its nominal value. Single faults $f_x$ and $f_y$ are assigned a random fault site in layers $SF2$ and $SF1$, respectively. Multiple fault $f_z$ is initialized with $4$ random fault sites anywhere in the network. Fault $f_x$ is injected to the network as the first single-fault fault round using the $cmpn.inject$ function. Then, faults $f_y$ and $f_z$ are included in a second five-fault fault round using the $cmpn.then\_inject$ function. The function $cmpn.inject$ always adds the faults to the current fault round, while $cmpn.then\_inject$ adds a new fault round. The next line calls the $cmpn.run$ function that takes as argument a reference to the complete testing set $test\_set$ and performs the preparation and execution stages of the FI campaign, as described in Sections \ref{sec:framework_FI_campaign, preparation_stage} and \ref{sec:framework_FI_campaign_execution_stage}. By default, $cmpn.run$ makes use of all available optimization options described in Section \ref{sec:framework_optimizations}, but the user can opt to use late start or early stop. Once the FI campaign is over, the details of the FI experiment, i.e., fault rounds, and the results, i.e., matrix $A_f^L$, classification accuracy for each fault round, etc., are extracted as a FI campaign data object using the function $cmpn.export$ and stored using the $cmpn.save$ function. Depending on the preferred results visualization, the user can choose among a set of plotting functions, as it will be demonstrated in Section \ref{sec:demonstrations}. The FI campaign data object is fed to the plotting functions to visualize the results. In this example, the results are plotted using the $bar$ function. This FI campaign terminates by calling the $cmpn.eject$ function, which removes all fault rounds and re-initializes the network to perform a new FI campaign if desired, allowing for unlimited reuse of the campaign object $cmpn$.
A second FI campaign is then executed using the function $cmpn.inject\_complete$, which completes the inject functions family. $cmpn.inject\_complete$ creates as many fault rounds with single faults as the number of processing elements in the specified layer. In this example, we perform a bit-flip in the MSB of an 8-bit integer representation for all synapses in layer $SF2$. This second FI campaign is executed with the $cmpn.run$ function and the results are visualized in a form of a heat map by calling the $heat$ function.

\section{Results}
\label{sec:results}

\subsection{Case Studies}
\label{sec:case_studies}

\textit{SpikeFI} is demonstrated on two convolutional SNNs trained to classify the N-MNIST \cite{OJCT15} and IBM's DVS128 Gesture \cite{ATBM17} datasets. These SNNs are modelled and trained in SLAYER, and are included in the \textit{demo} package of the \textit{SpikeFI} framework in GitHub. They use rate coding, i.e., the winning class is selected after the neuron at the output layer which is triggered the most producing the highest number of spikes.

\begin{figure}[t]
    \centering
\includegraphics[width=1.0\columnwidth]{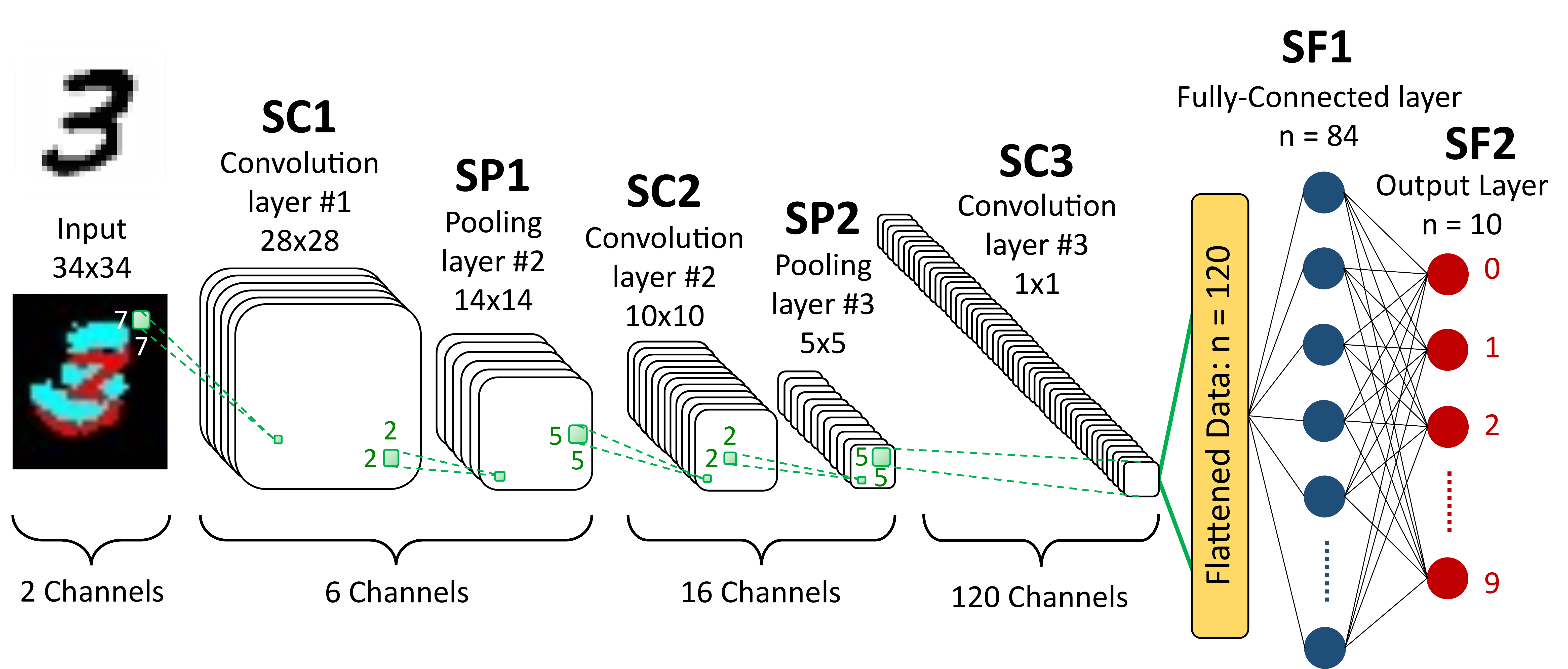}
    \caption{N-MNIST SNN.}
    \label{fig:architecture_nmnist}
\end{figure}

The N-MNIST dataset is a neuromorphic, i.e., spiking, version of the MNIST dataset, which comprises images of handwritten arithmetic digits in gray-scale format \cite{OJCT15}. It consists of 70000 sample images that are generated from the saccadic motion of a DVS in front of the original images in the MNIST dataset. The samples in the N-MNIST dataset have a duration of 300 ms. The dataset is split into a training set of 60000 samples and a testing set of 10000 samples. The SNN architecture, shown in Fig. \ref{fig:architecture_nmnist}, is a spiking version of the LeNet-5 architecture \cite{LBBH98}. It consists of 3 convolutional layers with 2 2x2 sum-pooling layers in between them and 2 fully-connected layers at the end for the final decision of the network. The classification accuracy on the testing set is 97.8\%.

\begin{figure}[t]
    \centering
    \includegraphics[width=1.0\columnwidth]{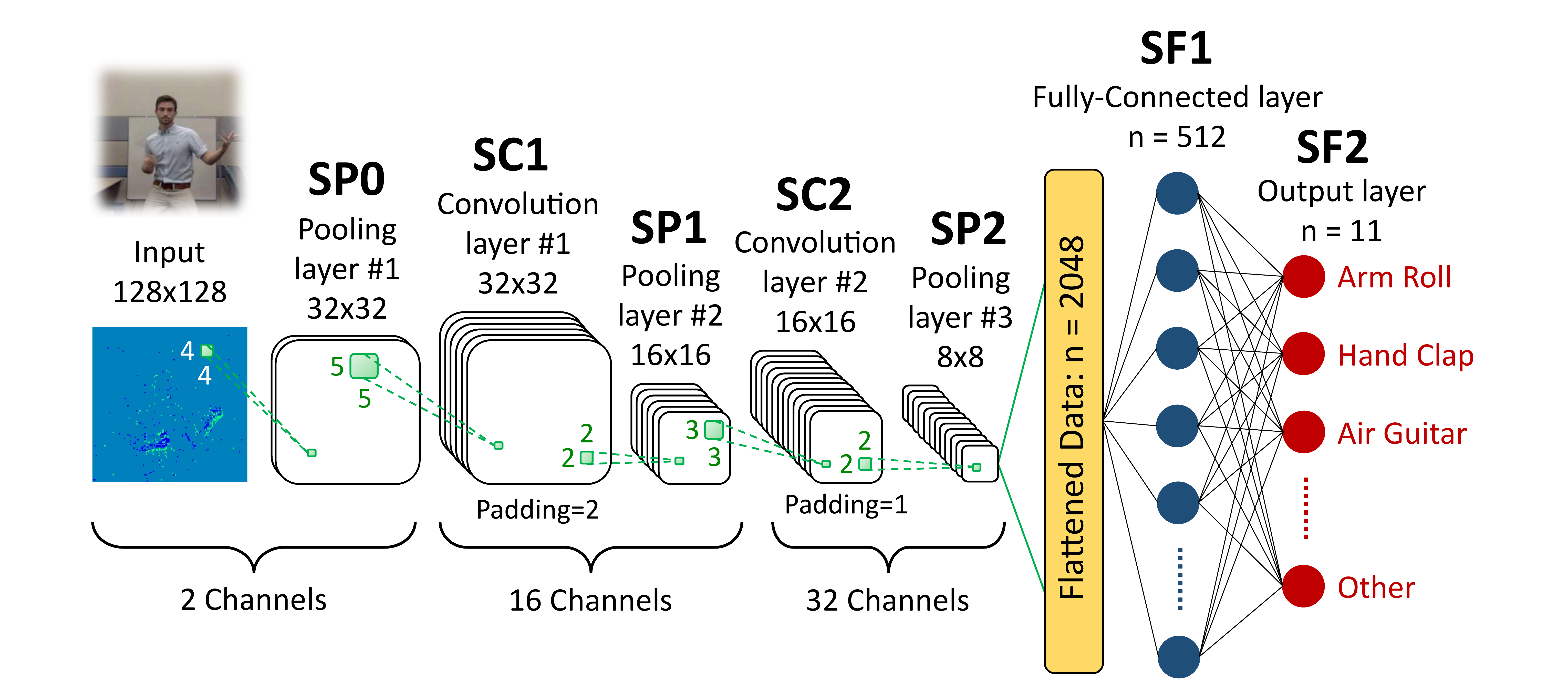}
    \caption{IBM DVS128 Gesture SNN.}
    \label{fig:architecture_gesture}
\end{figure}

The IBM's DVS128 Gesture dataset consists of 29 individuals performing 11 hand and arm gestures in front of a DVS, such as hand waving and air guitar, under 3 different lighting conditions \cite{ATBM17}. In total, the dataset comprises 1342 samples of duration 6 s, which is trimmed to 1.5 s to speedup simulation. The designed SNN, shown in Fig. \ref{fig:architecture_gesture}, is an adaptation of
the network proposed in \cite{ATBM17}. It starts with a 4x4 sum-pooling layer to reduce the big size of the input samples. Next, there are 2 convolutional layers followed by a 2x2 sum-pooling layer each. The architecture is concluded with 2 fully-connected layers. The network performs with an 86.4\% accuracy on the testing set, which is acceptable considering the shortened samples of the dataset and the shallower architecture compared to the architecture in \cite{ATBM17}.

\subsection{Optimization speedups}\label{sec:optimization_speedups}

Herein, we use the N-MNIST SNN as a benchmark for quantifying the speedup improvements when using the different optimizations.

For a fair comparison, all the FI experiments below were executed one at a time on the same system configuration composed of an Intel Xeon\textsuperscript{\textregistered} W-2133 CPU and a NVIDIA Quadro\textsuperscript{\textregistered} RTX 4000 GPU, with the system being reserved for the experiment.

\subsubsection{Ordering of nested  for loops}

\begin{figure}[t]
\centering
    \includegraphics[width=1\columnwidth] {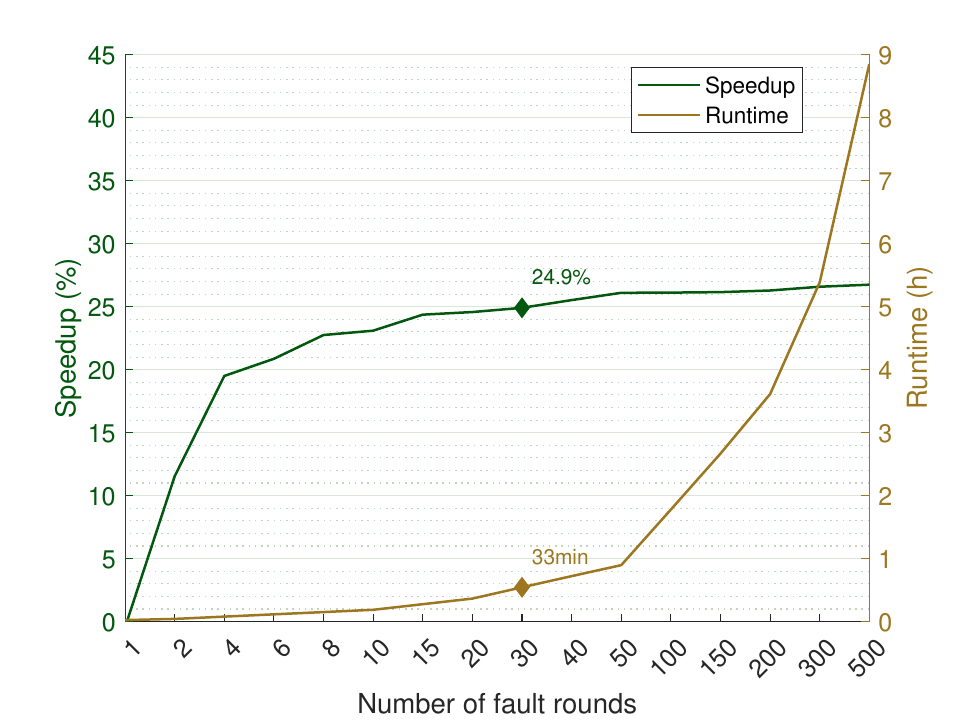}
    \caption{Speedup and runtime when using fault rounds in the inner \textit{for} loop.}
    \label{fig:speedup_O1}
\end{figure}

First, we show that placing the fault rounds in the inner \textit{for} loop as opposed to the outer \textit{for} loop speeds up the analysis. For this purpose, we perform successive FI campaigns with a batch size of 1 and with increasing number of fault rounds, and we measure the speedup as well as the total runtime. Each fault round is composed of a single random dead neuron fault, and the faults in the successive fault rounds are accumulated, i.e., one fault round contains all faults of the previous fault round plus a new random one. For this experiment, the early stop and late start optimizations are disabled. The result is shown in Fig. \ref{fig:speedup_O1}. As it can be seen, the speedup increases exponentially with the number of fault rounds, with the speedup slowing down after 10 fault rounds and converging to around 27\%. The runtime increases linearly with the number of fault rounds (note that the scale of the x axis is not linear). The convergence occurs when the runtime starts dominating, overshadowing the benefit from the reduced data transfers to the GPU. Based on this result and as discussed in Section \ref{sec:ordering_for_loops}, \textit{SpikeFI} places the fault rounds in the inner \textit{for} loop.

\subsubsection{Late start and early stop}

\begin{figure}[t]
\centering
    \includegraphics[width=1\columnwidth] {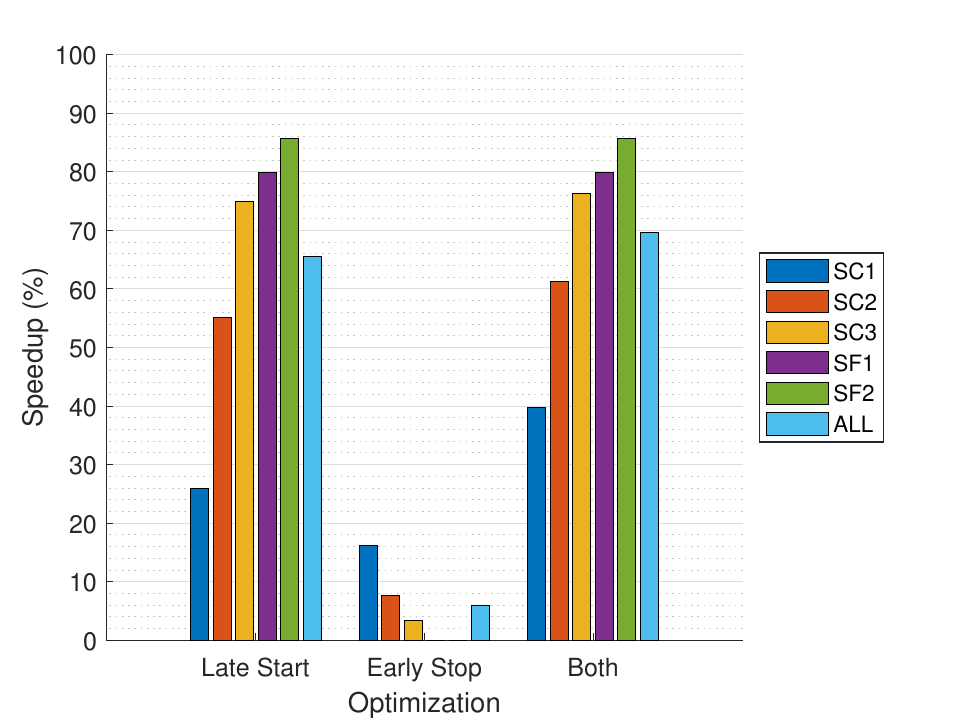}
    \caption{\centering Speedup using the late start and early stop optimizations.}
    \label{fig:speedup_Ox_vs_layer}
\end{figure}

To demonstrate the speedups offered by late start and early stop, we use a FI experiment with a batch size of 1 and 30 fault rounds, with each fault round containing a single random dead neuron fault. As marked in Fig. \ref{fig:speedup_O1}, such an experiment takes up approximately 33 min. The FI experiment is repeated layer-wise for all 5 layers by concentrating all 30 faults in one layer. For the last layer that has only 10 neurons, we triplicate each fault round. The FI experiment is also performed network-wise by distributing the fault rounds equally across the 5 layers, i.e., placing 6 faults per layer. 

Fig. \ref{fig:speedup_Ox_vs_layer} shows the speedup for these layer-wise and network-wise FI experiments when using late start and early stop optimizations as stand-alone and combined. For the early stop we use $\epsilon=0$. As expected, we observe that the speedup offered by late start improves as the leftmost faulty layer moves to the right. In contrast, early stop is more effective as the rightmost faulty layer moves to the left. As the fault model is dead neuron faults, late start can offer speedup even for faults in the first layer (see Section \ref{sec:late_start}).

Another observation is that late start offers significantly greater speedups compared to early stop. For the late start the speedup ranges from 26\% to 86\% moving from layer 1 to layer 5, while for early stop the speedup is less spectacular. It ranges from 16\% to 3\% moving from layer 1 to layer 3, while it vanishes for layers 4 and 5. For the network-wise FI experiment, late start offers a speedup of 66\%, whereas the speedup for early stop is a modest 6\%. Combining both the speedup reaches 70\%. The reason behind late start being more effective than early stop is that early stop is activated only when a fault round is benign and also requires the evaluation of $\| B^l \|_1$ whose time can counterbalance the average speedup benefit. In contrast, late start is applied to any type of fault round and is activated instantaneously. Early stop can offer a significant speedup for the initial layers when a large percentage of faults are benign.

\begin{figure}[t]
\centering
    \includegraphics[width=1\columnwidth] {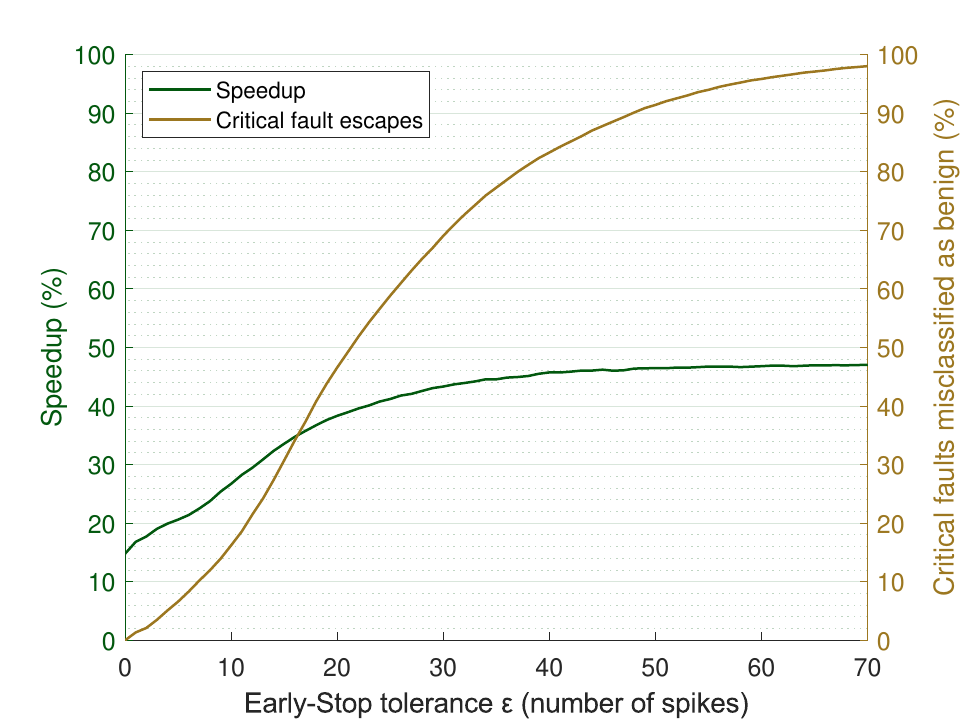}
    \caption{Speedup and critical fault escape rate when using early stop with tolerance $\epsilon>0$.}
    \label{fig:speedup_es_tolerance}
\end{figure}

Fig. \ref{fig:speedup_es_tolerance} depicts the effect of using $\epsilon>0$ in early stop. For this experiment, the FI campaign is performed on the first layer, containing as many fault rounds as the number of neurons in this layer, each with a single dead neuron fault. We observe that as $\epsilon$ increases, the percentage of critical faults misclassified as benign increases. For this FI experiment, improving further the speedup is not possible without misclassifying critical faults. For $\epsilon=1$, the speedup is around 15\% while the misclassification moves away from zero. Therefore, early stop with tolerance $\epsilon>0$ should be used with caution, as it could lead to masking the effect of critical faults.

\subsubsection{Batched inference}

\begin{figure}[t]
\centering
    \includegraphics[width=1\columnwidth] {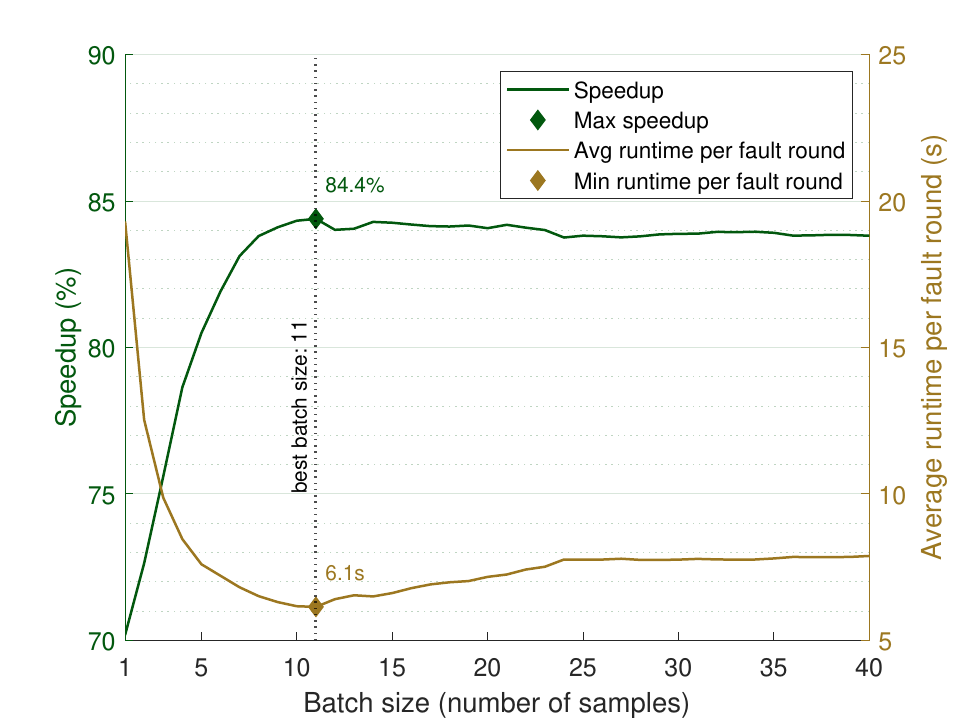}
    \caption{Speedup using batched inference.}
    \label{fig:speedup_batch_size}
\end{figure}

Finally, Fig. \ref{fig:speedup_batch_size} shows the speedup and average runtime per fault round as a function of batch size. The FI experiment is composed of 30 fault rounds of a single random dead neuron fault, distributed equally across the 5 layers. The late start and early stop options are turned on, using $\epsilon=0$ for the early stop. The baseline speedup for batch size 1 is 70\%, as shown in the last column of Fig. \ref{fig:speedup_Ox_vs_layer}, where the exact same FI experiment is performed. As it can be seen from Fig. \ref{fig:speedup_batch_size}, the speedup increases by a maximum of approximately $15\%$ for batch size 11, converging to around $85\%$ for larger batch sizes, while the average runtime per fault round reaches a minimum of $6.1$ s at this point. The reason behind this initial speedup improvement is that, on one hand, PyTorch is optimized to calculate more efficiently the output for many input samples simultaneously and, on the other hand, by increasing the batch size we reduce the data transfer time to and from the GPU. However, at some specific batch size, the bottleneck of the GPU communication is reached, saturating the speedup improvement.

\subsection{Demonstrations}\label{sec:demonstrations}

\subsubsection{Neuron hard faults}
\label{sec:demonstrations_hard_neuron}

\begin{figure}[t]
\centering
\begin{subfigure}[b]{1\columnwidth}
        \centering
        \includegraphics[width=1\columnwidth] {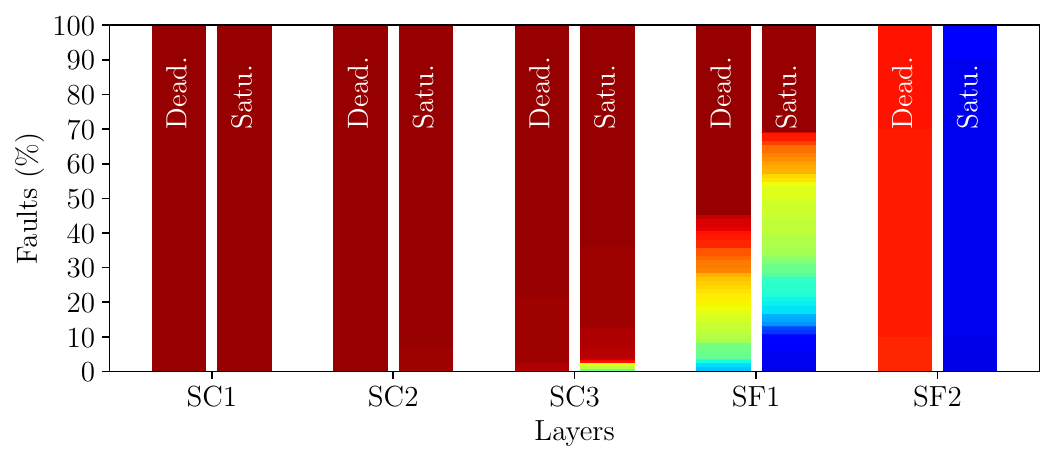}
        \caption{N-MNIST SNN.}
        
        \includegraphics[width=1\columnwidth] {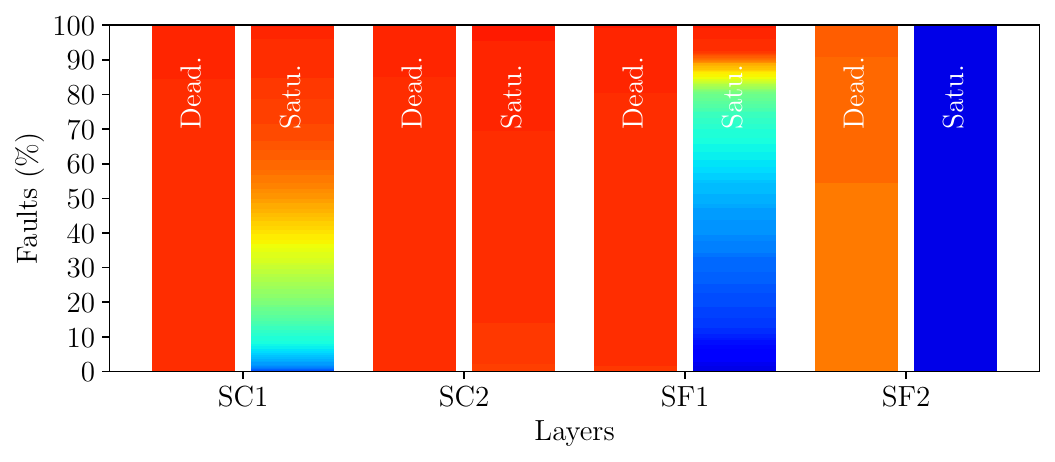}
        \caption{IBM DVS128 Gesture SNN.}
    \end{subfigure}

    \begin{subfigure}[b]{1\columnwidth}  
        \centering 
        \includegraphics[width=\textwidth]{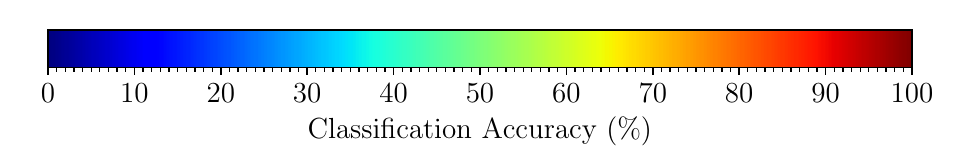}
    \end{subfigure}
    
    \caption{Resiliency analysis for  dead and saturated neuron faults.}
    \label{fig:neuron_dead_satu_bar}
\end{figure}

In the first experiment, we inject single hard neuron faults considering all neurons. We ran two separate FI campaigns for dead and saturated neuron faults. 

Fig. \ref{fig:neuron_dead_satu_bar} shows a possible visualization of the results using comparative bar plots. The x-axis shows the different layers of the network and for each layer there are two bars, one for the dead and one for the saturated neuron faults. Note that pooling layers were excluded from the analysis since their functionality is to aggregate regions of spikes of their previous layers and do not contain any spiking neurons. A bar is separated into chunks of different colors, each corresponding to a specific classification accuracy according to the color shading shown at the bottom of Fig. \ref{fig:neuron_dead_satu_bar}. The height of the chunk projected on the y-axis shows the percentage of neurons in this layer which when exhibit this type of fault the classification accuracy drops to the value indicated by the color of the chunk.

A first observation from Fig. \ref{fig:neuron_dead_satu_bar} is that saturated faults have a far stronger impact on the classification accuracy compared to dead faults. At the output layer, a saturated neuron always wins the race, thus samples from all classes except the one corresponding to the winning neuron are always misclassified. In the case of a dead neuron, an input with class label corresponding to this neuron is always mislassified, while samples from other classes are not affected. Taking the N-MINST SNN as an example, a saturated neuron at the output layer causes the accuracy to plummet to a value of 10\% on average, while a dead neuron reduces the accuracy by 10\% on average. The fact that saturated faults are more lethal than dead faults is also evident in the SF1 layer of the two networks. The IBM DVS128 Gesture SNN is impacted also in the SC1 layer, while the N-MNIST SNN is insensitive to faults in the first two layers and in the third layer only a 2\% of neurons are critical.

\subsubsection{Neuron parametric faults}\label{sec:demo_neuron_parametric_faults}

\begin{figure*}[t]
    \centering
    \begin{subfigure}[b]{0.32\textwidth}
        \centering
        \includegraphics[width=\textwidth]{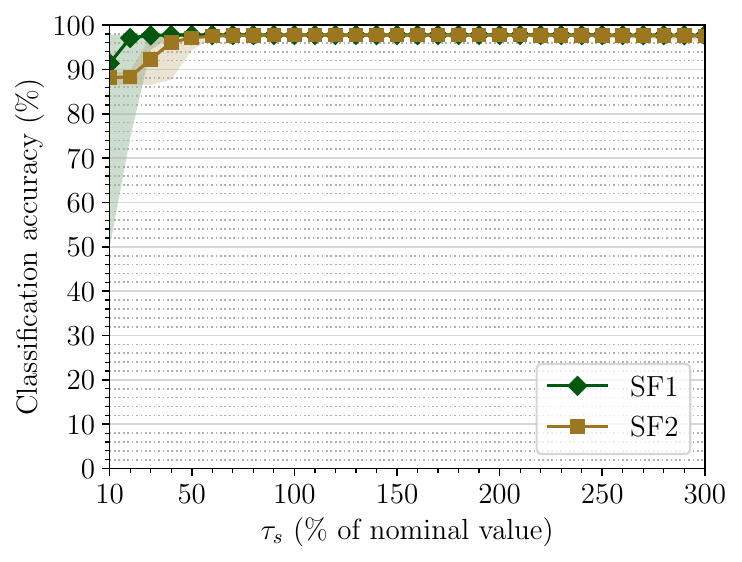}
        \caption{N-MNIST SNN: accuracy vs. $\tau_s$.}
        \label{fig:neuron_param_tausr_nmnist}
    \end{subfigure}
    \hfill
    \begin{subfigure}[b]{0.32\textwidth}
        \centering
        \includegraphics[width=\textwidth]{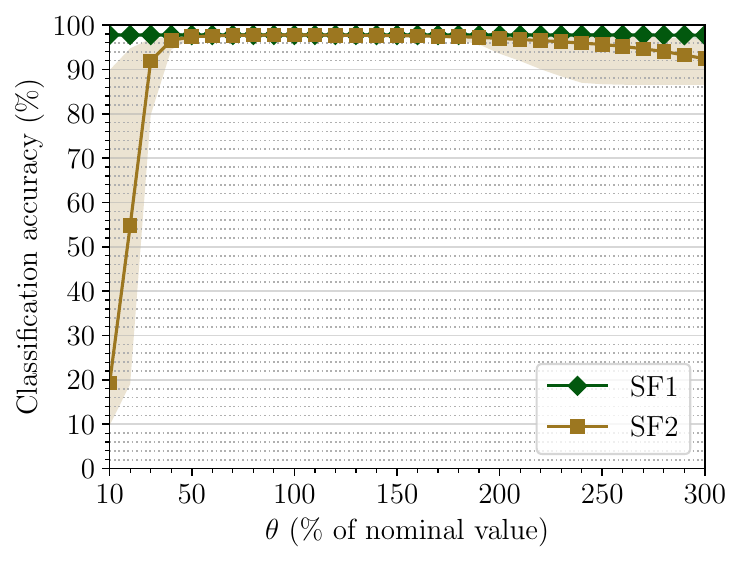}
        \caption{N-MNIST SNN: accuracy vs. $\theta$.}
        \label{fig:neuron_param_theta_nmnist}
    \end{subfigure}
    \hfill
    \begin{subfigure}[b]{0.32\textwidth}
        \centering
        \includegraphics[width=\textwidth]{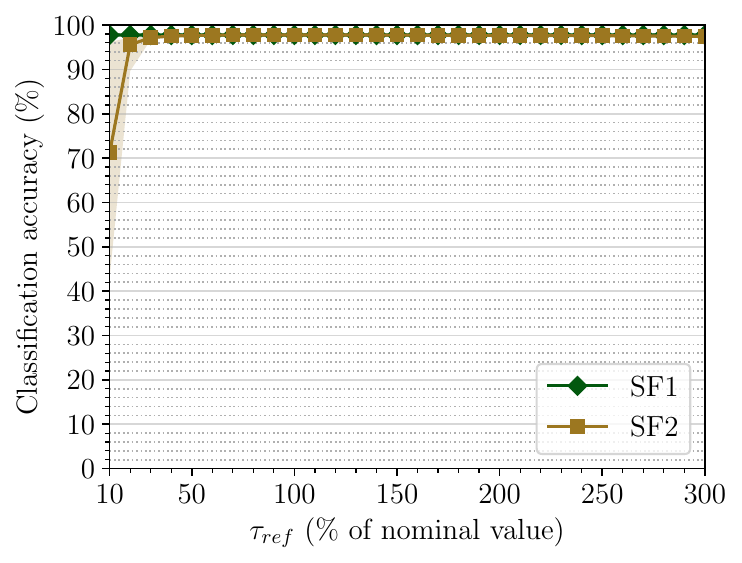}
        \caption{N-MNIST SNN: accuracy vs. $\tau_{ref}$.}
        \label{fig:neuron_param_tauref_nmnist}
    \end{subfigure}
    \vskip\baselineskip

    \begin{subfigure}[b]{0.32\textwidth}
        \centering
        \includegraphics[width=\textwidth]{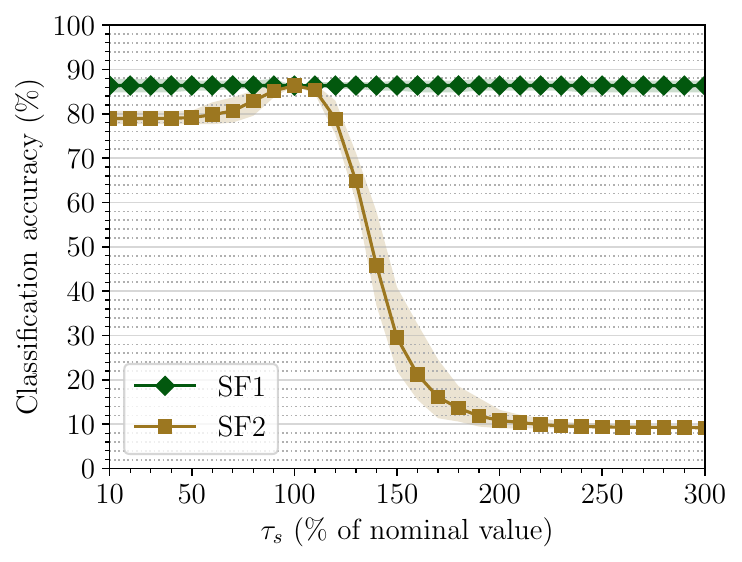}
        \caption{IBM DVS128 Gesture SNN: accuracy vs. $\tau_s$.}
        \label{fig:neuron_param_tausr_gesture}
    \end{subfigure}
    \hfill
    \begin{subfigure}[b]{0.32\textwidth}
        \centering
        \includegraphics[width=\textwidth]{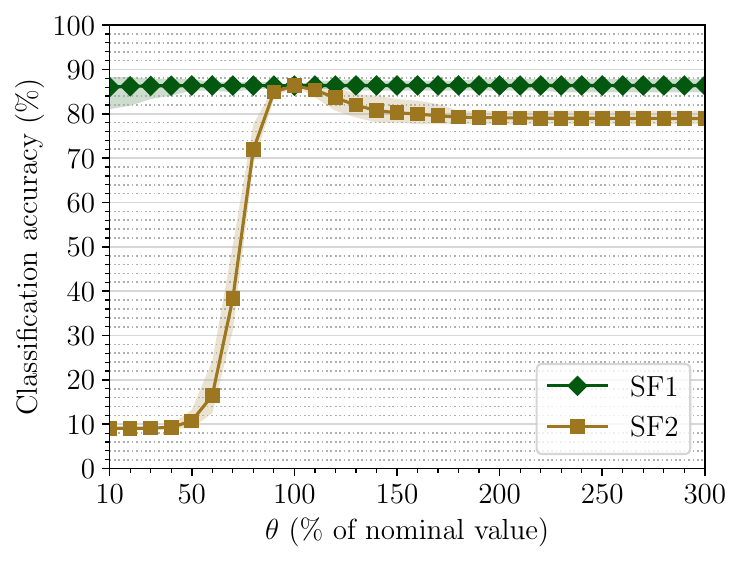}
        \caption{IBM DVS128 Gesture SNN: accuracy vs. $\theta$.}
        \label{fig:neuron_param_theta_gesture}
    \end{subfigure}
    \hfill
    \begin{subfigure}[b]{0.32\textwidth}
        \centering
        \includegraphics[width=\textwidth]{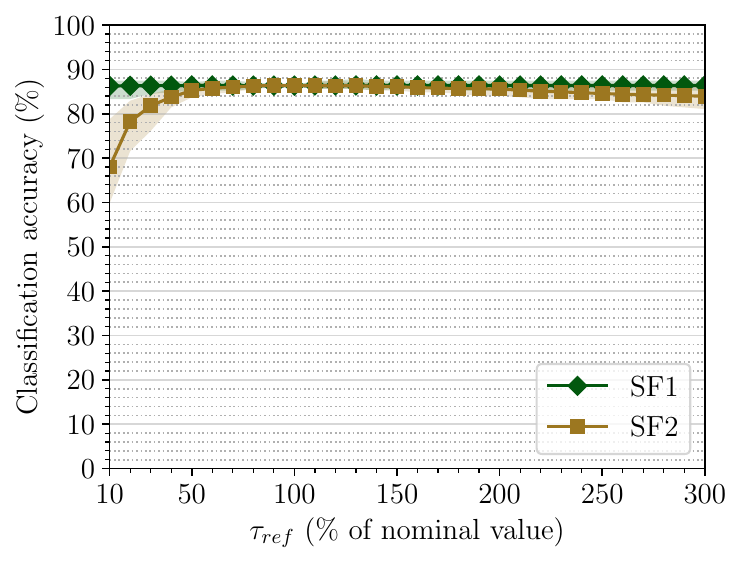}
        \caption{IBM DVS128 Gesture SNN: accuracy vs. $\tau_{ref}$.}
        \label{fig:neuron_param_tauref_gesture}
    \end{subfigure}

    \caption{Resiliency analysis for neuron parametric faults.}
    \label{fig:neuron_param_plot}
\end{figure*}

The effect of neuron parametric faults on the classification accuracy is shown in Fig. \ref{fig:neuron_param_plot}. For a given layer, we vary the $\tau_s$, $\tau_{ref}$, and $\theta$ parameters in the SRM for one neuron at a time. The main curve represents in the y-axis the average classification accuracy observed across all neurons of the layer as a function of the parameter deviation in the x-axis expressed in \% of the nominal value, i.e., 100\% corresponds to zero deviation. The colored region surrounding the curve demonstrates the minimum and maximum classification accuracy. Neuron parametric faults were found to have a noticeable impact only for the output layer, thus in Fig. \ref{fig:neuron_param_plot} we show only the results for the output and last hidden layer.

The N-MNIST SNN shows a high resilience even at the output layer. The classification accuracy starts degrading when $\tau_s$, $\theta$, and $\tau_{ref}$ are reduced at 40\%, 40\%, and 20\%, respectively, while positive deviations have practically no effect as accuracy degradation starts being noticeable only for $\theta$ when it increases beyond 200\%. In contrast, the SNN DVS128 Gesture SNN shows vulnerability even for small $\tau_s$ and $\theta$ fluctuations, while it shows a high degree of resiliency for $\tau_{ref}$.

As mentioned in Section \ref{sec:neuron_parametric_faults}, increasing $\tau_s$, decreasing $\theta$, or decreasing $\tau_{ref}$ makes the neuron spike more easily. At the extreme, this direction of deviation may make the neuron saturate, which, as we observed in Section \ref{sec:demonstrations_hard_neuron}, is far more fatal than a dead fault. This behavior can be observed in Fig. \ref{fig:neuron_param_plot}. Extreme positive deviation of $\tau_s$ for the IBM DVS128 Gesture SNN has an effect equivalent to neuron saturation as the accuracy drops to 9.09\%, i.e., only 1 out of 11 classes is predicted correctly. Similarly, extreme negative deviation of $\theta$ is equivalent to neuron saturation in both networks. 

\subsubsection{Synapse faults}\label{sec:demo_synpases}

\begin{figure}[t]
    \centering
    \begin{subfigure}[b]{1\columnwidth}
        \centering
        \includegraphics[width=1\textwidth]{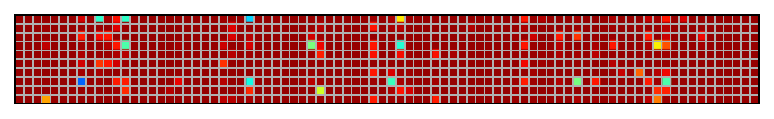}
        \\
        \includegraphics[width=0.985\textwidth]{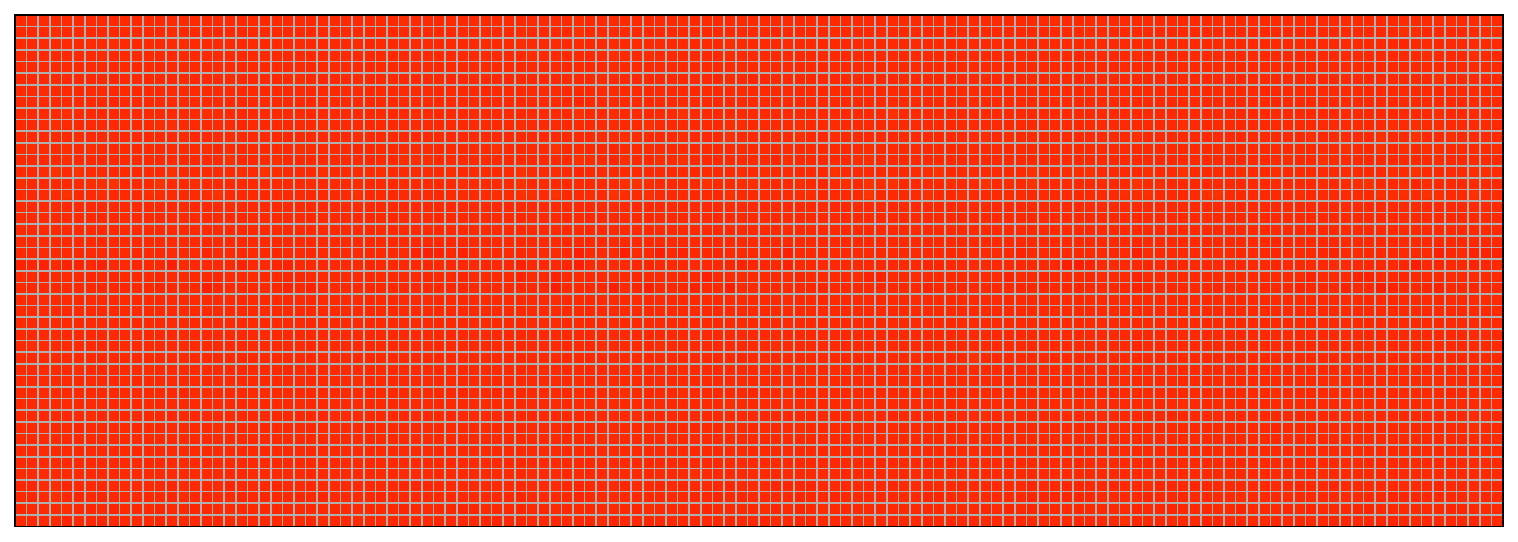}
        \caption{Dead synapse fault.}
        \label{fig:synapse_dead_heat}
    \end{subfigure}
    \vskip\baselineskip
    
    \begin{subfigure}[b]{1\columnwidth}
        \centering
        \includegraphics[width=1\textwidth]{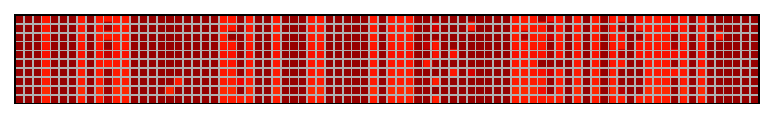}
        \\
        \includegraphics[width=0.985\textwidth]{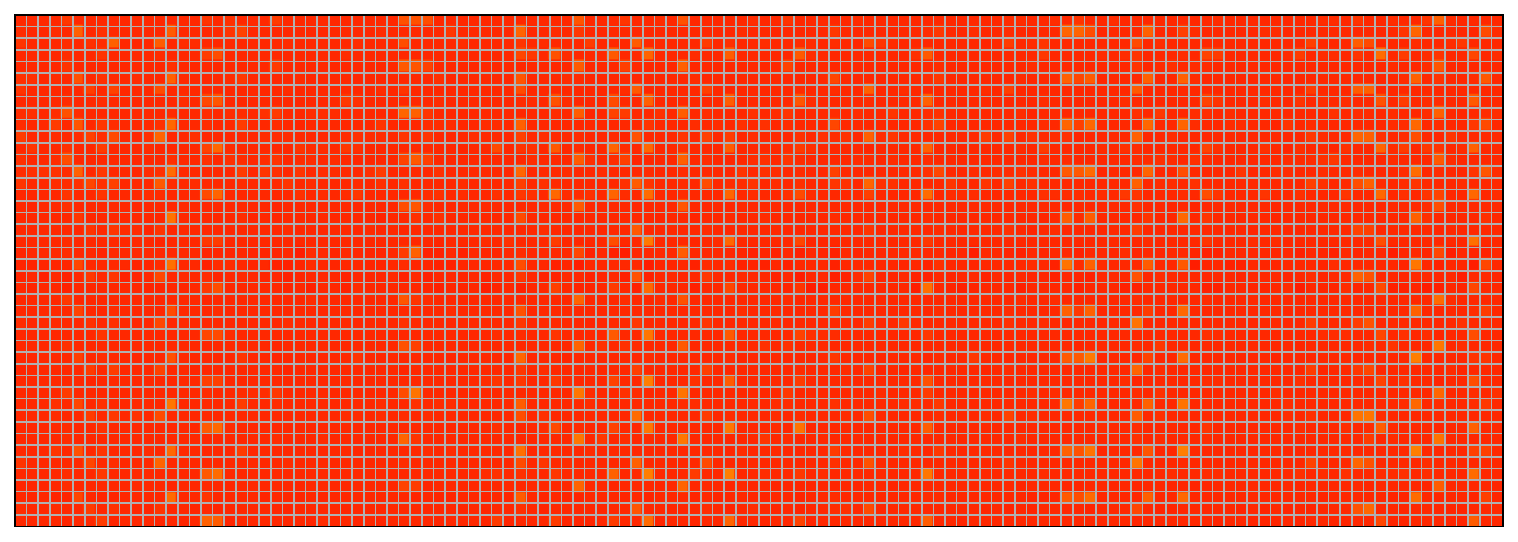}
        \caption{Negatively saturated synapse fault.}
        \label{fig:synapse_satu-_heat}
    \end{subfigure}
    \vskip\baselineskip
    
    \begin{subfigure}[b]{1\columnwidth}
        \centering
        \includegraphics[width=1\textwidth]{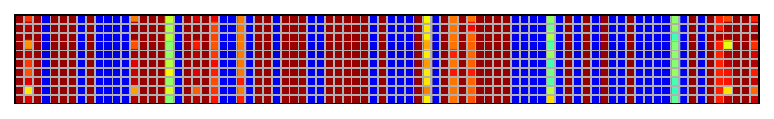}
        \\
        \includegraphics[width=0.985\textwidth]{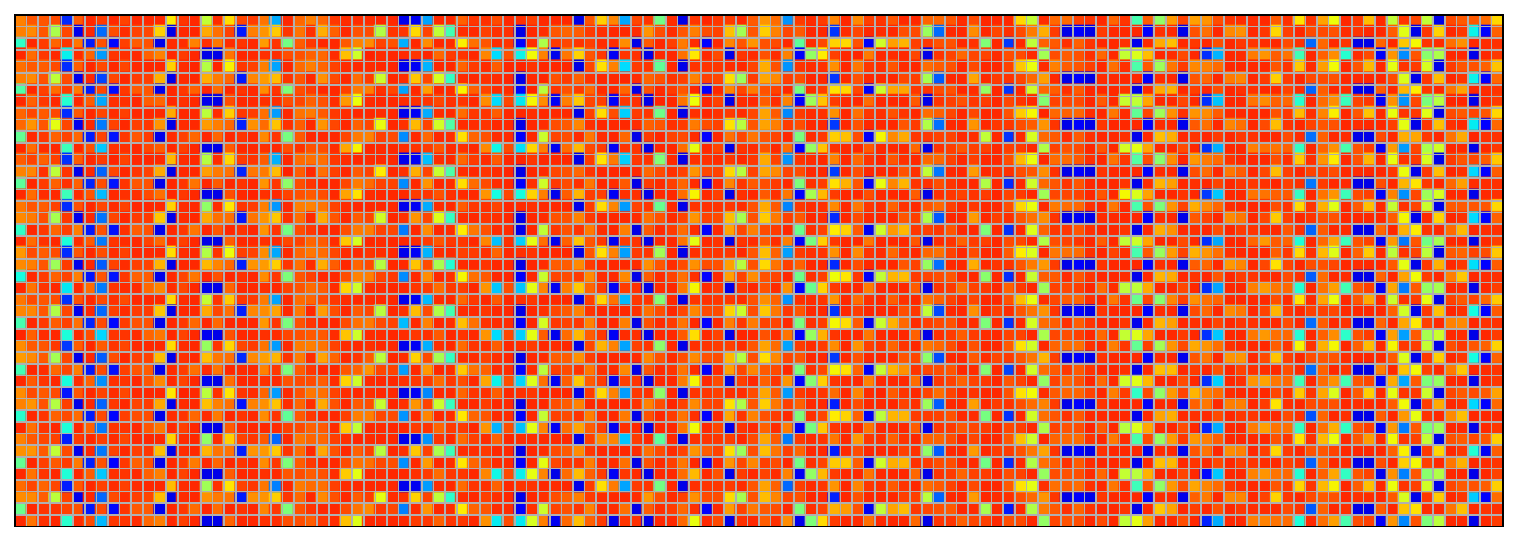}
        \caption{Positively saturated synapse fault.}
        \label{fig:synapse_satu+_heat}
    \end{subfigure}

    \begin{subfigure}[b]{1\columnwidth}  
        \centering 
        \includegraphics[width=\textwidth]{Figs/FI_Experiments/colormap.pdf}
    \end{subfigure}
    
    \caption{Resiliency analysis for synapse faults in synapses connecting the last two layers. In each sub-figure the result for the N-MNIST SNN is shown at the top and for the IBM DVS128 Gesture SNN at the bottom.}
    \label{fig:synapse_dead_satu_heat}
\end{figure}

\begin{figure*}
    \centering
    \begin{subfigure}[b]{0.24\columnwidth}
        \centering
        \includegraphics[width=\textwidth]{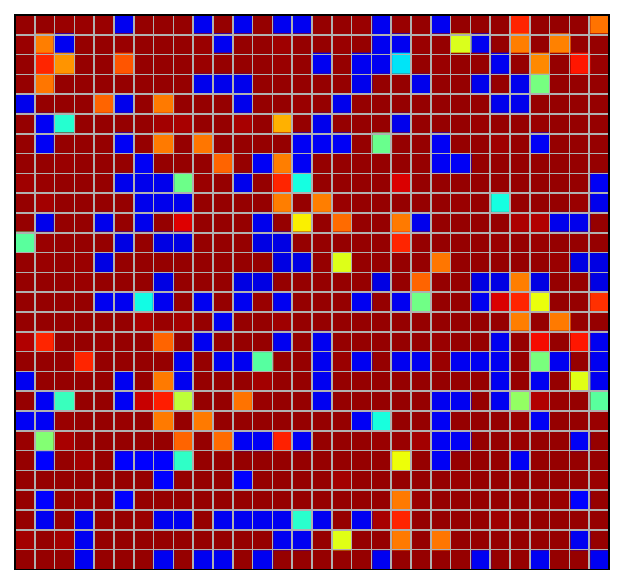}
        \caption{Bit 0 (LSB).}    
    \end{subfigure}
    \hfill
    \begin{subfigure}[b]{0.24\columnwidth}
        \centering
        \includegraphics[width=\textwidth]{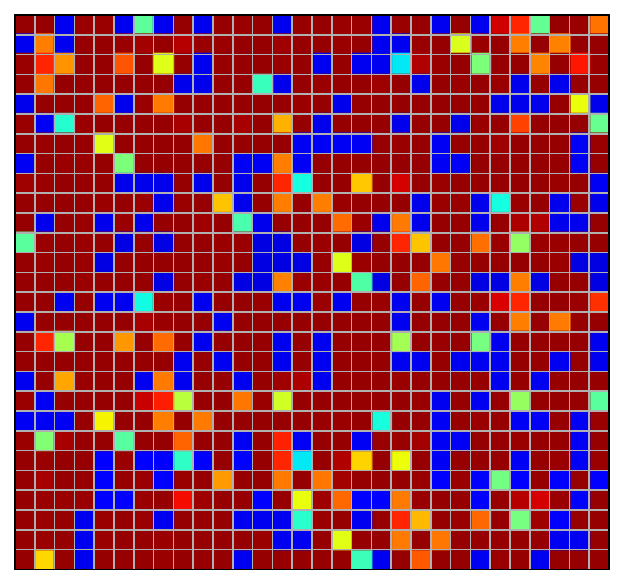}
        \caption{Bit 1.}    
    \end{subfigure}
    \hfill
    \begin{subfigure}[b]{0.24\columnwidth}
        \centering
        \includegraphics[width=\textwidth]{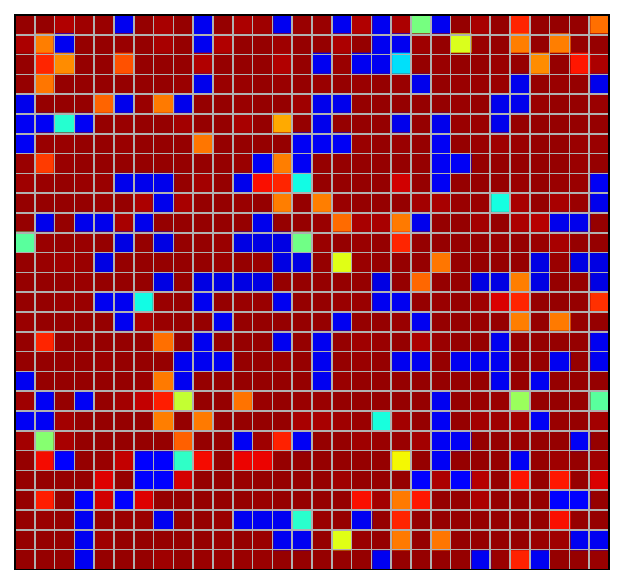}
        \caption{Bit 2.}    
    \end{subfigure}
    \hfill
    \begin{subfigure}[b]{0.24\columnwidth}
        \centering
        \includegraphics[width=\textwidth]{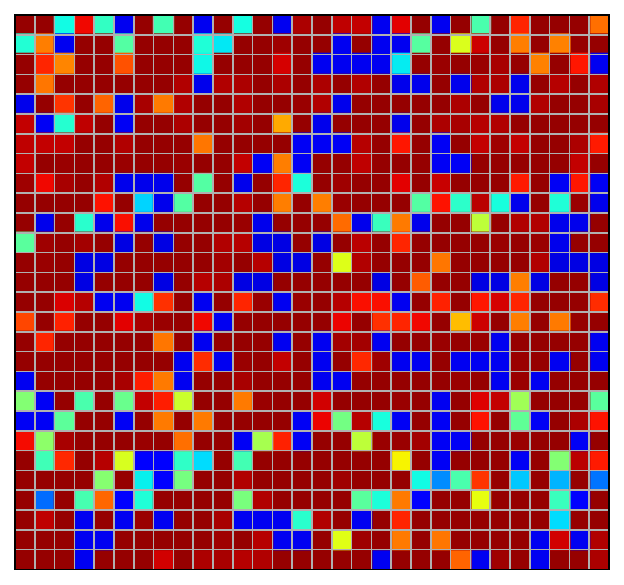}
        \caption{Bit 3.}    
    \end{subfigure}
    \hfill
    \begin{subfigure}[b]{0.24\columnwidth}
        \centering
        \includegraphics[width=\textwidth]{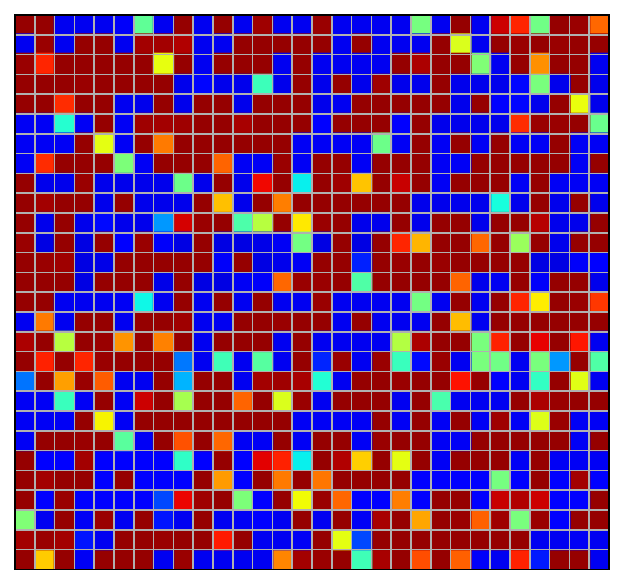}
        \caption{Bit 4.}    
    \end{subfigure}
    \hfill
    \begin{subfigure}[b]{0.24\columnwidth}
        \centering
        \includegraphics[width=\textwidth]{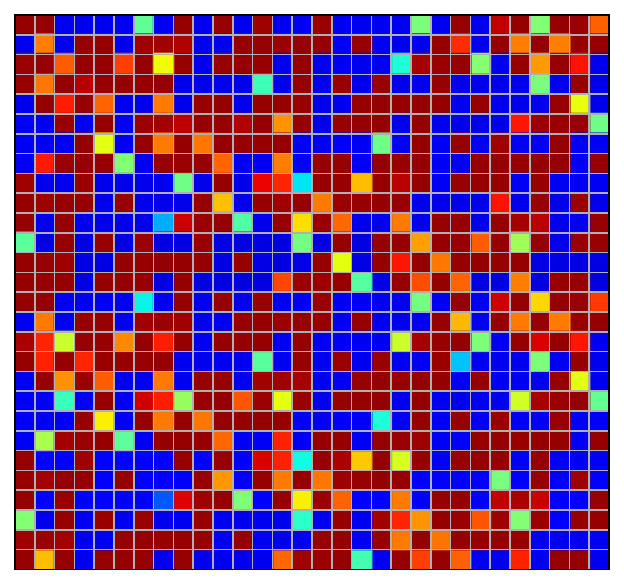}
        \caption{Bit 5.}    
    \end{subfigure}
    \hfill
    \begin{subfigure}[b]{0.24\columnwidth}
        \centering
        \includegraphics[width=\textwidth]{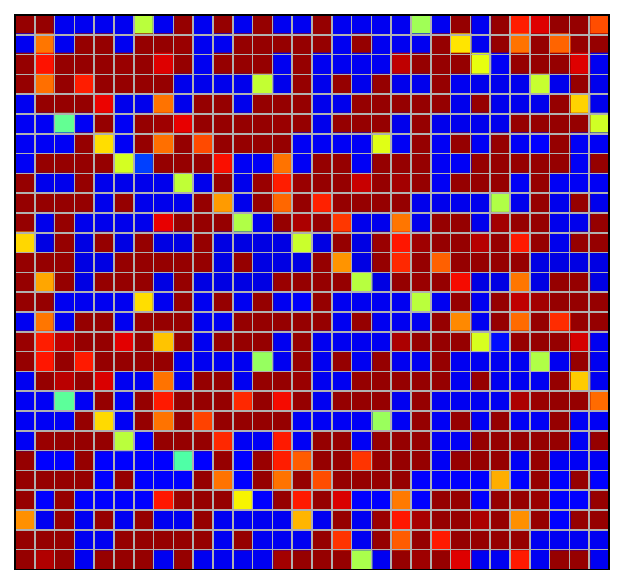}
        \caption{Bit 6.}    
    \end{subfigure}
    \hfill
    \begin{subfigure}[b]{0.24\columnwidth}
        \centering
        \includegraphics[width=\textwidth]{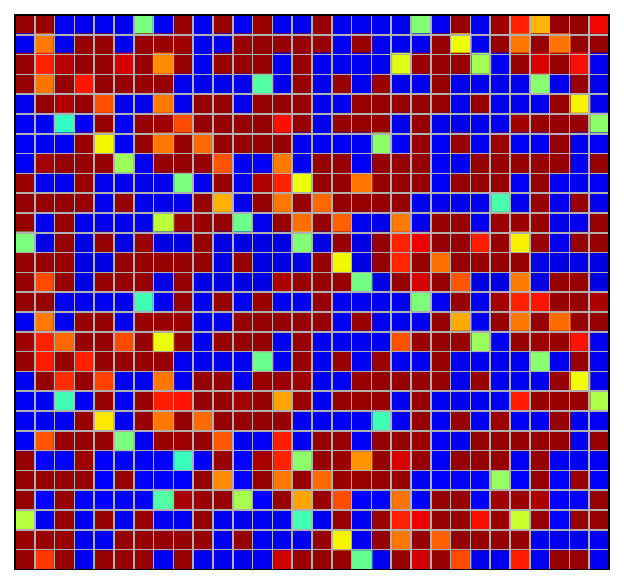}
        \caption{Bit 7 (MSB).}    
    \end{subfigure}
    \vskip\baselineskip

    \begin{subfigure}[b]{0.49\columnwidth}
        \centering
        \includegraphics[width=\textwidth]{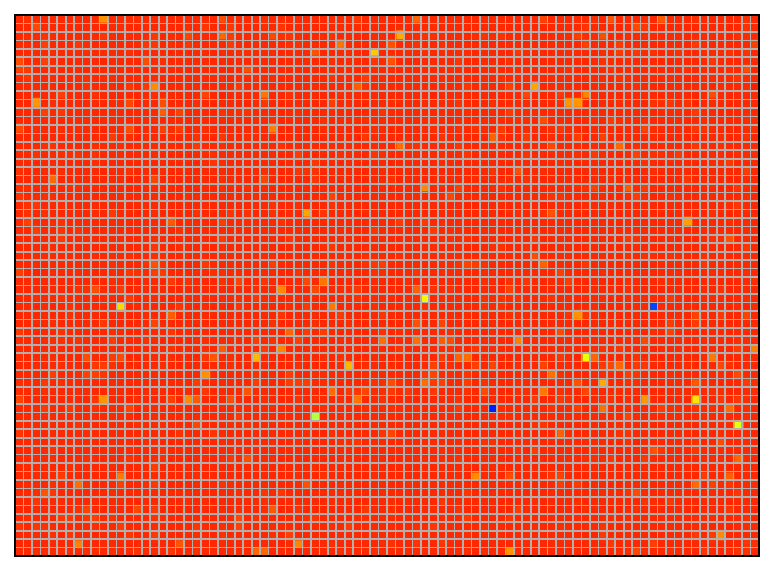}
        \caption{Bit 0 (LSB).}    
    \end{subfigure}
    \hfill
    \begin{subfigure}[b]{0.49\columnwidth}
        \centering
        \includegraphics[width=\textwidth]{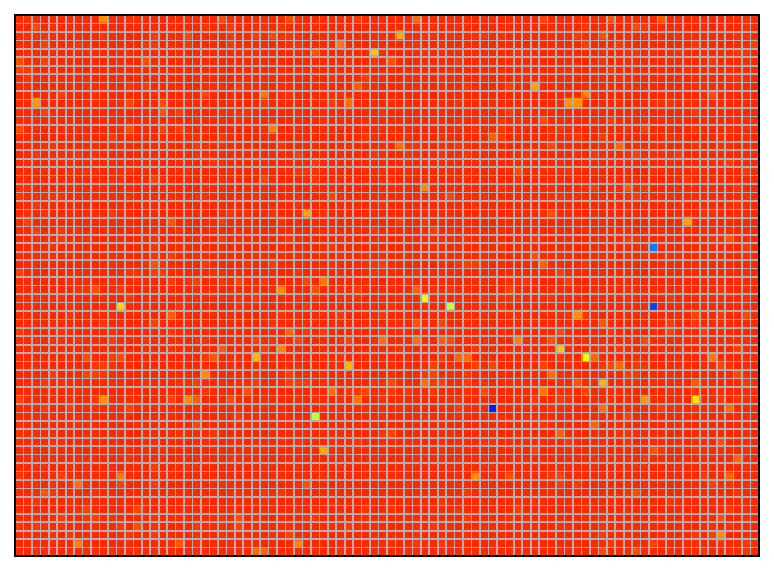}
        \caption{Bit 1.}    
    \end{subfigure}
    \hfill
    \begin{subfigure}[b]{0.49\columnwidth}
        \centering
        \includegraphics[width=\textwidth]{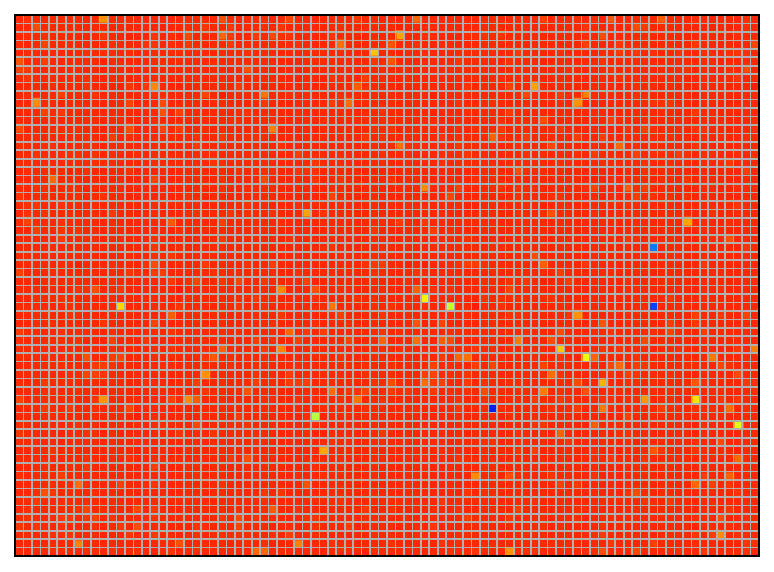}
        \caption{Bit 2.}    
    \end{subfigure}
    \hfill
    \begin{subfigure}[b]{0.49\columnwidth}
        \centering
        \includegraphics[width=\textwidth]{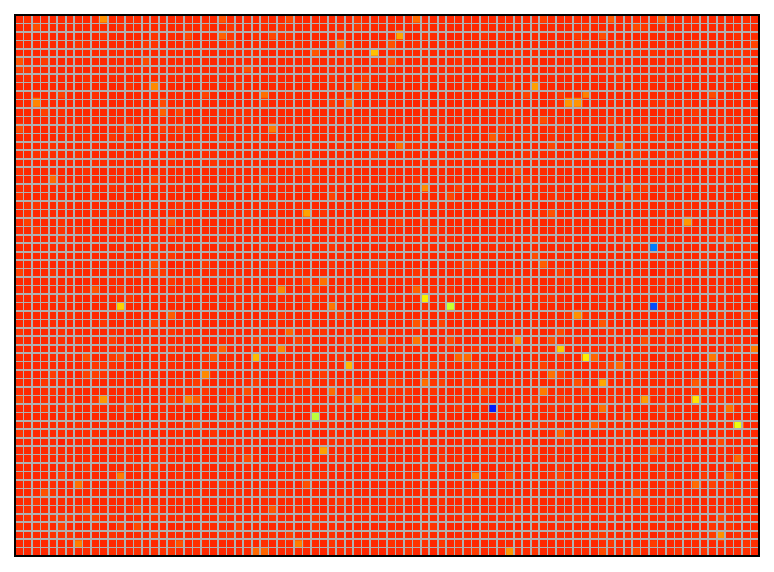}
        \caption{Bit 3.}    
    \end{subfigure}
    \vskip\baselineskip
    
    \begin{subfigure}[b]{0.49\columnwidth}
        \centering
        \includegraphics[width=\textwidth]{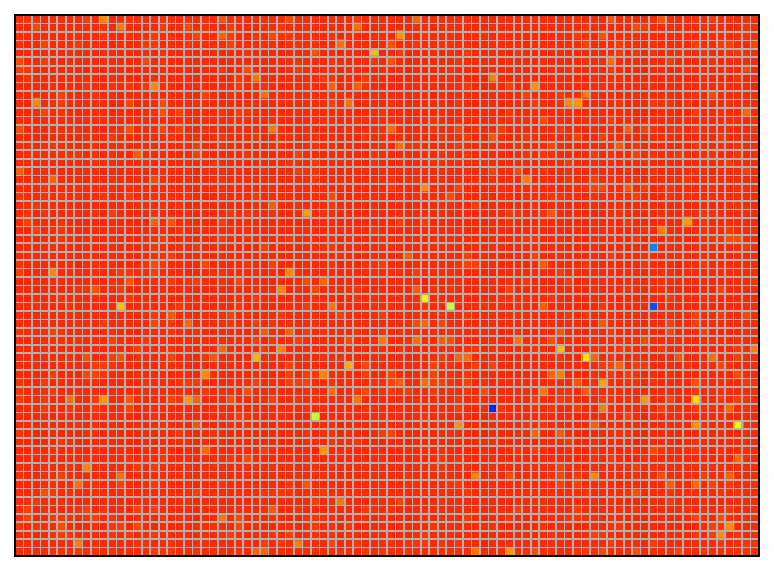}
        \caption{Bit 4.}    
    \end{subfigure}
    \hfill
    \begin{subfigure}[b]{0.49\columnwidth}  
        \centering 
        \includegraphics[width=\textwidth]{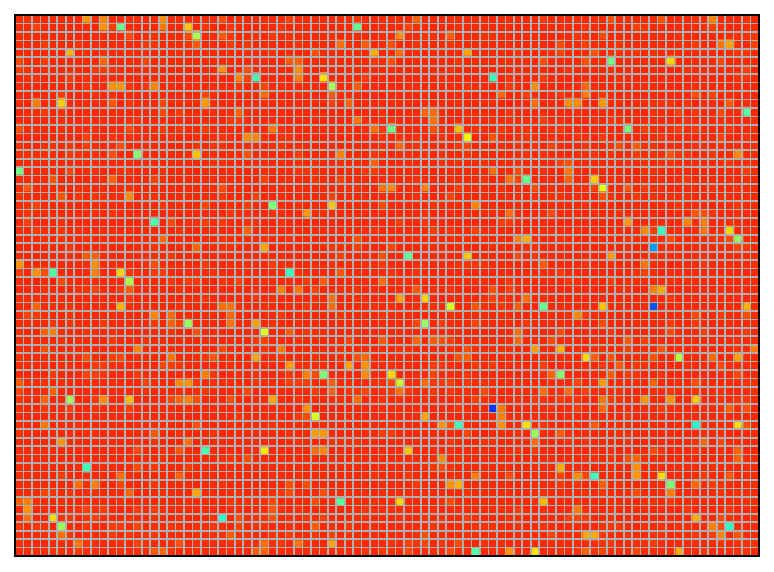}
        \caption{Bit 5.}
    \end{subfigure}
    \hfill
    \begin{subfigure}[b]{0.49\columnwidth}
        \centering
        \includegraphics[width=\textwidth]{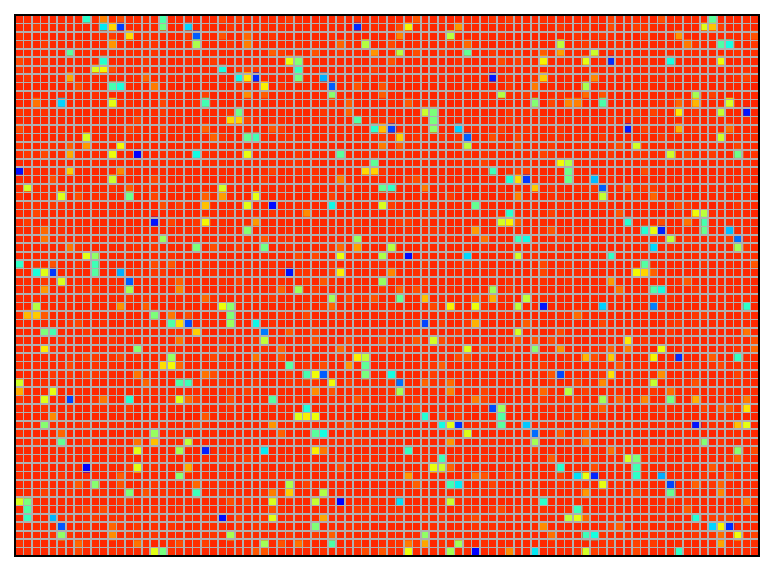}
        \caption{Bit 6.}    
    \end{subfigure}
    \hfill
    \begin{subfigure}[b]{0.49\columnwidth}  
        \centering 
        \includegraphics[width=\textwidth]{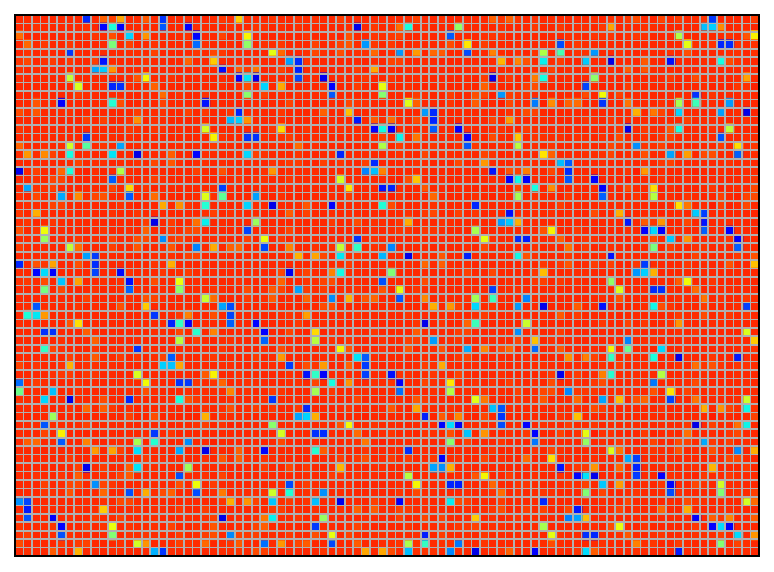}
        \caption{ Bit 7 (MSB).}
    \end{subfigure}
    
     \begin{subfigure}[b]{1\columnwidth}  
        \centering 
        \includegraphics[width=\textwidth]{Figs/FI_Experiments/colormap.pdf}
    \end{subfigure}
    
    \caption{Resiliency analysis for bit-flips in synapses connecting the last two layers for the N-MNIST SNN (top heat maps) and IBM DVS128 Gesture SNN (bottom heat maps).}
    \label{fig:synapse_bitflip_heat}
\end{figure*}

We consider four different synapse faults, namely dead, negatively and positively saturated, and bit-flipped synapses. The negative and positive saturation values were set to $-10$ and $+10$, respectively, which are extreme considering the synaptic weight distribution after training. For the bit-flipped synapse faults we consider that the hardware accelerator uses an 8-bit integer data format. For each synapse fault model, we performed an exhaustive FI campaign with single synapse faults covering all synapses connecting the last hidden layer to the output layer. The result is illustrated in the form of heat maps in Figs. \ref{fig:synapse_dead_satu_heat} and \ref{fig:synapse_bitflip_heat}. Each square in the heat map corresponds to a unique synaptic connection and the square's color represents the network's resultant classification accuracy when the synapse fault is introduced. The framework outputs these heat maps such that the pre-synaptic neurons are placed in the x-axis and the post-synaptic neurons are placed in the y-axis. As the number of neurons can be very high, the framework offers the possibility to re-shape the area of the heat map for illustration purposes.

From Figs. \ref{fig:synapse_dead_heat} and \ref{fig:synapse_satu-_heat}, we observe that only a few synapses can affect the classification accuracy if they become dead or negatively saturated, and the drop in the classification accuracy is in most cases small. A dead synapse fault zeros the spikes passing from the synapse, while a negative saturated synapse fault converts the spikes to large negative spikes. In both cases, the membrane potential of the post-synaptic neuron reduces and at the extreme the neuron never fires behaving like a dead neuron. In contrast, positive saturated synapse faults can have a greater impact, as shown in Fig. \ref{fig:synapse_satu+_heat}. This is because they increase the spikes' strength to the point where the post-synaptic neuron fires easily, behaving at the extreme like a saturated neuron.

From Fig. \ref{fig:synapse_bitflip_heat}, we observe that the effect on the classification accuracy increases with the bit position, with the most significant bits (MSBs) being the most critical. We observe also that the N-MNIST SNN is the most vulnerable to this synapse fault type, as even the least significant bit (LSB) is critical for many synapses. 

\subsubsection{Runtime}

\begin{table}[t]
    \centering
    \caption{Statistics of the FI campaigns in Sections \ref{sec:demonstrations_hard_neuron}, \ref{sec:demo_neuron_parametric_faults} and \ref{sec:demo_synpases}.}
    \scriptsize
    \begin{tabular}{ccccc}
    \toprule

        Network
        & Fault type
        & \begin{tabular}{@{}c@{}}
            Fault \\ rounds (\#)
        \end{tabular}
        & \begin{tabular}{@{}c@{}}
            Total \\ runtime (sec)
        \end{tabular}
        & \begin{tabular}{@{}c@{}}
            Runtime \\ per fault (sec)
        \end{tabular} \\ \toprule

         \multirow{3}*[-0.2em]{\rotatebox{90}{N-MNIST}}
         & Neuron hard  & 13036 & 163982 & 12.58 \\ \cmidrule{2-5}
         & Neuron parametric &  8460 &  32290 &  3.82 \\ \cmidrule{2-5}
         & Synapse &  9240 &  15404 &  1.67 \\ \midrule
         
         \multirow{3}*[-0.75em]{\rotatebox{90}{\parbox{0.8cm}{IBM \\DVS128 \\Gesture}}}
         & Neuron hard  & 50198 & 66989 & 1.33 \\ \cmidrule{2-5}
         & Neuron parametric  & 47070 & 18066 & 0.34 \\ \cmidrule{2-5}
         & Synapse & 61952 &  4768 & 0.08 \\

    \bottomrule
    \end{tabular}
    \label{tab:demo_fi_experiments}
\end{table}

Table \ref{tab:demo_fi_experiments} summarizes the statistics of the FI campaigns in the above Sections \ref{sec:demonstrations_hard_neuron}, \ref{sec:demo_neuron_parametric_faults} and \ref{sec:demo_synpases}. It shows the total number of fault rounds, total runtime, and average runtime per fault round per network and per fault type. Overall, these experiments involved 189956 fault rounds and took around $3.5$ days to complete. Synapse fault simulation presents the smallest runtime, while for neuron parametric faults the runtime is one order of magnitude smaller than neuron hard faults.

\subsubsection{Training in the presence of faults}

\begin{figure}[t]
\centering
    \includegraphics[width=1\columnwidth] {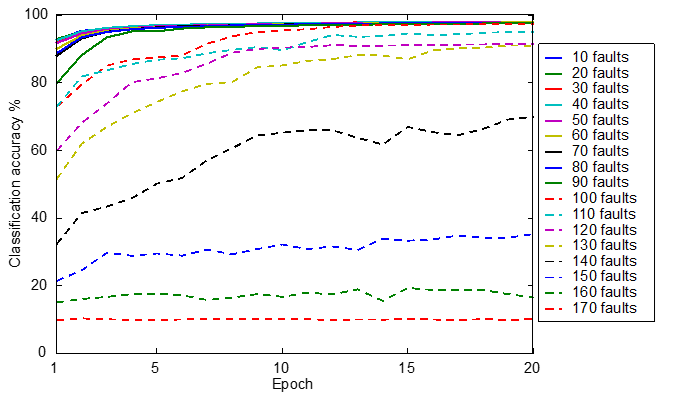}
    \caption{Learning curves of the training of the N-MNIST SNN in the presence of multiple faults of various types.}
    \label{fig:fault_aware_training}
\end{figure}

Herein, \textit{SpikeFI} is used to assess the ability of a SNN to learn in the presence of faults. The FI experiment consists of several cumulative fault rounds, where in each fault round 10 new faults are added with respect to the previous one. The fault sites within a fault round are randomly assigned across the network excluding the output layer. The fault target can be any neuron or synapse, and a fault function from those defined in the fault model library is randomly assigned. For every fault round, the faults are injected into the network and then training is performed. Fig. \ref{fig:fault_aware_training} shows the learning curves for the N-MNIST SNN. Each learning curve corresponds to a separate fault round and results in a new instance of the SNN. As it can be seen, the SNN is capable of learning around and withstanding multiple faults. The nominal fault-free accuracy is reached for up to 100 faults, although the learning rate slows down as the number of faults increases. The classification accuracy starts dropping after 100 simultaneous faults occur in the network, with the drop increasing with the number of faults. This experiment shows that SNN hardware accelerators used for training present a high degree of passive tolerance to faults existing prior to training, and that re-training when faults occur is a workable active fault tolerance approach at the expense of bringing the network temporarily offline.

\section{Conclusions}
\label{sec:conclusions}

We described \textit{SpikeFI}, an open-source GPU-accelerated FI framework for SNNs. \textit{SpikeFI} is built on top of the popular SLAYER framework used for training SNNs. It has built-in a library of mainstream neuron and synapse fault types modeled onto the SRM that can be extended and customized by the user if desired. \textit{SpikeFI} enables highly flexible FI experiments for any arbitrary SNN model. Each FI experiment is composed of independent fault rounds, where, in turn, each fault round can be composed of single or multiple faults of different types. The fault duration can be permanent or transient and the fault site can be specified or can be randomly assigned layer-wise or network-wise. \textit{SpikeFI} can be used to train a SNN with faults to make it robust to the presence of faults, study the effect on the training accuracy for faults manifesting during a long training process, or study the effect on the inference accuracy for faults occurring post-training so as to pinpoint critical faults and develop smart and low-cost test and fault tolerance techniques. \textit{SpikeFI} features various speed-up optimization tricks and various results visualization functions compatible with all fault types. \textit{SpikeFI} was demonstrated on two SNNs designed for the N-MNIST and IBM DVS128 Gesture datasets, which are the two most common benchmark datasets within the neuromorphic community.

\section*{Acknowledgments}
This work was funded by the ANR RE-TRUSTING project under Grant N$^{\mbox{\scriptsize o}}$ ANR-21-CE24-0015-03 and by the European Network of Excellence dAIEDGE under Grant Agreement N$^{\mbox{\scriptsize o}}$ 101120726. The work of T. Spyrou was supported by the Sorbonne Center for Artificial Intelligence (SCAI) through Fellowship.

\bibliographystyle{IEEE}

\end{document}